\documentclass{article}

\PassOptionsToPackage{numbers, compress}{natbib}

\usepackage[final]{neurips_2021}

\usepackage[utf8]{inputenc} 
\usepackage[T1]{fontenc}    
\usepackage{hyperref}       
\usepackage{url}            
\usepackage{booktabs}       
\usepackage{amsfonts}       
\usepackage{nicefrac}       
\usepackage{microtype}      
\usepackage{xcolor}         
\usepackage{todonotes}
\usepackage{amsmath}
\usepackage{amsthm}
\usepackage{paralist}
\usepackage{float}
\usepackage{multirow}
\usepackage{pifont}
\usepackage{booktabs}
\usepackage[normalem]{ulem}
\useunder{\uline}{\ul}{}
\newcommand{\cmark}{\ding{51}}%
\newcommand{\xmark}{\ding{55}}%

\newtheorem{example}{Example}
\newtheorem{definition}{Definition}

\title{Benchmarking the Combinatorial Generalizability of \\ Complex Query Answering on Knowledge Graphs}

\author{%
  Zihao Wang\thanks{Equal Contribution} \\
  Department of CSE\\
  HKUST\\
  \texttt{zwanggc@cse.ust.hk} \\
   \And
  Hang Yin$^*$\\
  Department of Mathematical Sciences\\
  Tsinghua University \\
  \texttt{h-yin20@mails.tsinghua.edu.cn} \\
  \And
  Yangqiu Song\\
  Department of CSE, HKUST\\
  Peng Cheng Laboratory, Shenzhen, China\\
  \texttt{yqsong@cse.ust.hk}
}

\begin{document}

\maketitle

\begin{abstract}
Complex Query Answering (CQA) is an important reasoning task on knowledge graphs. Current CQA learning models have been shown to 
be able to generalize from atomic operators to more complex formulas, which can be regarded as the combinatorial generalizability. In this paper, we present EFO-1-QA, a new dataset to benchmark the combinatorial generalizability of CQA models by including 301 different queries types, which is 20 times larger than existing datasets. Besides, our benchmark, for the first time, provide a benchmark to evaluate and analyze the impact of different operators and normal forms by using (a) 7 choices of the operator systems and (b) 9 forms of complex queries. Specifically, we provide the detailed study of the combinatorial generalizability of two commonly used operators, i.e., projection and intersection, and justify the impact of the forms of queries given the canonical choice of operators. Our code and data can provide an effective pipeline to benchmark CQA models.
\end{abstract}

\section{Introduction}
Knowledge graphs, such as Freebase~\cite{freebase}, Yago~\cite{suchanek2007yago}, DBPedia~\cite{auer2007dbpedia}, and NELL~\cite{carlson2010toward} are graph-structured knowledge bases that can facilitate many fundamental AI-related tasks such as reasoning, question answering, and information retrieval~\cite{DBLP:conf/i-semantics/EhrlingerW16}.
Different from traditional well-defined ontologies, knowledge graphs often have the Open World Assumption (OWA), where the knowledge can be incomplete to support sound reasoning.
On the other hand, the graph-structured data naturally provide solutions to higher-order queries such as ``{\it the population of the largest city in Ohio State}.''


Given the OWA and scales of existing knowledge graphs, traditional ways of answering muti-hop queries can be difficult and time-consuming~\cite{DBLP:conf/nips/RenL20}.
Recently, several studies use learning algorithms to reason over the vector space to answer logical queries of complex types, e.g., queries with multiple projections~\cite{DBLP:conf/iclr/DasDZVDKSM18}, Existential Positive First-Order (EPFO) queries~\cite{DBLP:conf/nips/HamiltonBZJL18, DBLP:conf/iclr/RenHL20,DBLP:conf/iclr/ArakelyanDMC21}, and the so called first order queries, i.e., EPFO queries with the negation operator~\cite{DBLP:conf/nips/RenL20,DBLP:conf/nips/SunAB0C20,DBLP:journals/corr/abs-2103-00418}.
These tasks are usually called Complex Query Answering (CQA).
Unlike the traditional link predictors that only model entities and relations~\cite{DBLP:conf/nips/BordesUGWY13}, CQA models also consider logical connectives (operators) such as conjunction ($\land$), disjunction ($\lor$), and negation ($\lnot$) by parameterized operations~\cite{DBLP:conf/nips/HamiltonBZJL18,DBLP:conf/iclr/RenHL20,DBLP:conf/nips/RenL20} or non-parameterized operations such as logical t-norms~\cite{DBLP:conf/iclr/ArakelyanDMC21,DBLP:journals/corr/abs-2103-00418}.

One of the advantages of learning based methods is that the learned embeddings and parameterization in the vector space can generalize queries from 
atomic operations
to more complex queries.
It has been observed there are \emph{out-of-distribution generalization} phenomena of learning models~\cite{DBLP:conf/iclr/ArakelyanDMC21} on the Q2B dataset~\cite{DBLP:conf/iclr/RenHL20} (5 types to train but 4 unseen types to generalize) and the BetaE dataset~\cite{DBLP:conf/nips/RenL20} (10 types to train and 4 unseen types to generalize). 
This can be explained by the fact that complex queries are all composed by atomic operations such as projection, conjunction, disjunction, and negation. This idea evokes the \emph{combinatorial generalization}, that is, the model generalizes to novel combinations of already familiar elements~\cite{vankov2020training}.
However, compared to the huge combinatorial space of the complex queries (see Section~\ref{sec:cqa-over-kg} and \ref{sec:framework}), existing datasets~\cite{DBLP:conf/iclr/RenHL20,DBLP:conf/nips/RenL20} only contains queries from very few types, which might be insufficient for the investigation of the combinatorial generalization ability of learning models. Moreover, there is no agreement about how to present the complex queries by operators and normal forms. For example, some approaches treat the negation as the atomic operation~\cite{DBLP:conf/nips/RenL20,DBLP:journals/corr/abs-2103-00418} while others replace the negation by the set difference (intersection combined with negation)~\cite{DBLP:conf/nips/SunAB0C20,DBLP:conf/kdd/LiuDJZT21}. The impact of the representation of the complex query using learning algorithms is also unclear.

In this paper, we aim to benchmark the combinatorial generalizability of learning models for the CQA on knowledge graphs. We extend the scope from a few hand-crafted query types to the family of Existential First-Order queries with Single Free Variable (EFO-1) (see Section~\ref{sec:cqa-over-kg}) by providing a complete framework from the dataset construction to the model training and evaluation. Based on our framework, the combinatorial generalizability of CQA models that fully supports EFO-1 queries~\cite{DBLP:journals/corr/abs-2103-00418,DBLP:conf/nips/RenL20,DBLP:conf/kdd/LiuDJZT21} are evaluated and discussed.
Our contribution are in three-fold.

\begin{figure}
    \centering
    \includegraphics[width=.95\linewidth]{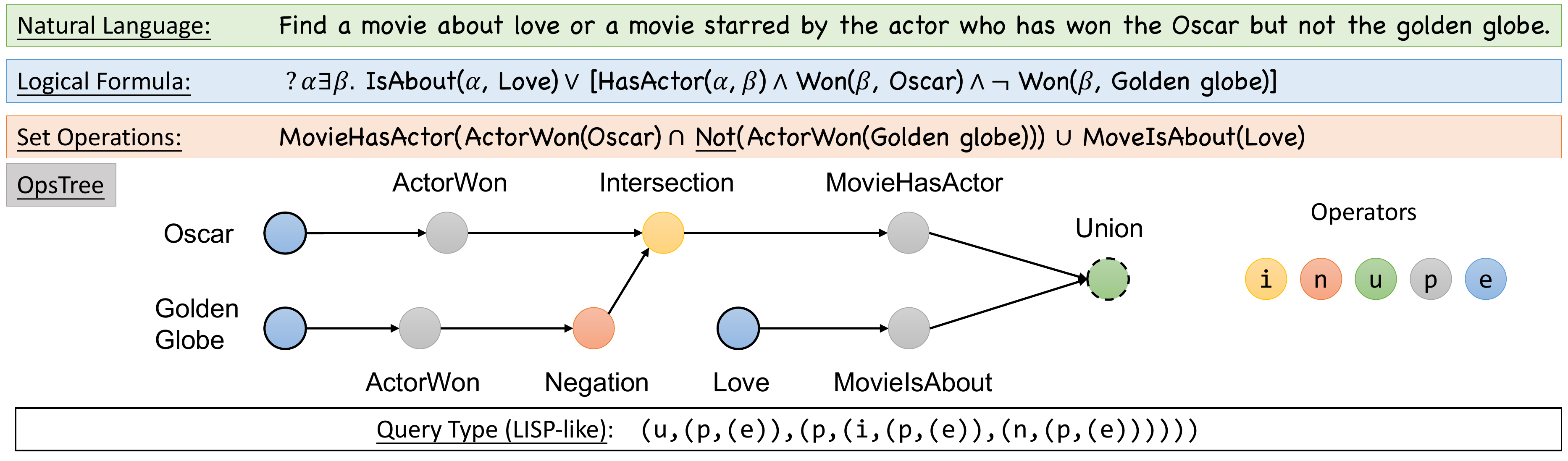}
    \vspace{-1em}
    \caption{An example of EFO-1 query. The same query are represented in natural language (which is only used for interpreting the query and we do not consider semantic parsing from natural language to logical forms), first-order logical formula, set operations, and OpsTree. The query type can be represented in the LISP-like grammar. 
    }
    \vspace{-1.1em}
    \label{fig:example}
\end{figure}

$\bullet\;${\bf Large-scale dataset of combinatorial queries.} We present the \textbf{EFO-1-QA} dataset to benchmark the combinatorial generalizability of CQA models. EFO-1-QA largely extends the scope of previous datasets by including 301 query types, which is 20 times larger than existing datasets. The evaluation results over three knowledge graphs show that the our set is generally harder than existing ones.

$\bullet\;${\bf Extendable framework.} We present a general framework for (1) iterating through the combinatorial space of EFO-1 query types, (2) converting queries to various normal forms with related operators, (3) sampling queries and their answer sets, and (4) training the CQA models and evaluating the CQA checkpoints. Our framework can be applied to generate new data as well as train and evaluate the models.

$\bullet\;${\bf New findings for normal forms, training, and generalization.} In our dataset, each query is transformed into at most 9 different forms that are related to 7 choices of operators. Therefore, for the first time, a deep analysis of normal forms are available in our benchmark. How the normal form affects the combinatorial generalization is discussed and new observations are revealed. Moreover, we also explore how training query types affect the generalization. We find that increasing training query types is not always beneficial for CQA tasks, which leads to another open problem about how to train the CQA models.


\section{Complex Queries on KG}~\label{sec:cqa-over-kg}
In this section, we introduce the Existential First Order Queries with Single Free Variable (EFO-1) on the knowledge graphs.
Here we give an intuitive example of EFO-1 queries and the related concepts in Figure~\ref{fig:example}. Compared to the query families considered in the existing works~\cite{DBLP:conf/nips/HamiltonBZJL18,DBLP:conf/iclr/RenHL20,DBLP:conf/nips/RenL20}, EFO-1 is a family of queries that are most general.
The formal definition and self-contained formal derivation of EFO-1 query family from first-order queries can be found in the Appendix~\ref{app:efo-1}.
Notably, the formal derivation of EFO-1 queries enables and guarantees the logical equivalent query representation in set operations and Operators Tree (OpsTree).
Specifically, the fomally derivied OpsTree is the composition of set functions including intersection, union, negation, projection, and entity anchors.
This presentation is also widely but informally introduced in existing CQA models~\cite{DBLP:conf/iclr/RenHL20,DBLP:conf/nips/RenL20,DBLP:conf/nips/SunAB0C20}.

We consider the EFO-1 queries at the \emph{abstract level} and the \emph{grounded level}. At the abstract level, the structure of a query is specified, but the projection or the entities are not given. At the grounded level, the projections and entities are instantiated (see Section~\ref{sec:framework} for how to ground the queries). We call queries without the instantiation \emph{query types}. When the query type is given, one can ground the projections and entities in a KG to obtain the specific EFO-1 query.

\begin{table}[t]
\centering
\caption{EFO-1 formula for 14 query types in BetaE dataset. The grammar of the EFO-1 formula are given in the Appendix~\ref{app:lisp-grammar}.}\label{tab:EFO-1-formula-of-beta}
\scriptsize
\begin{tabular}{llll}
\toprule
BetaE & EFO-1 formula           & BetaE & EFO-1 formula                 \\\midrule
1p               & \texttt{(p,(e))}                     & 3in              & \texttt{(i,(p,(e)),(i,(p,(e)),(n,(p,(e)))))}     \\
2p               & \texttt{(p,(p,(e)))}                 & inp              & \texttt{(p,(i,(p,(e)),(n,(p,(e)))))}         \\
3p               & \texttt{(p,(p,(p,(e))))}             & pin              & \texttt{(i,(p,(p,(e))),(n,(p,(e))))}         \\
2i               & \texttt{(i,(p,(e)),(p,(e)))}         & pni              & \texttt{(i,(n,(p,(p,(e)))),(p,(e)))}         \\
3i               & \texttt{(i,(i,(p,(e)),(p,(e))),(p,(e)))} & 2u-DNF           & \texttt{(u,(p,(e)),(p,(e)))}                 \\
ip               & \texttt{(p,(i,(p,(e)),(p,(e))))}     & up-DNF           & \texttt{(u,(p,(p,(e))),(p,(p,(e))))}         \\
pi               & \texttt{(i,(p,(p,(e))),(p,(e)))}     & 2u-DM            & \texttt{(n,(i,(n,(p,(e))),(n,(p,(e)))))}     \\
2in              & \texttt{(i,(p,(e)),(n,(p,(e))))}     & up-DM            & \texttt{(p,(n,(i,(n,(p,(e))),(n,(p,(e))))))}\\\bottomrule
\end{tabular}
\end{table}

\section{The Construction of EFO-1-QA Benchmark}~\label{sec:framework}

\begin{figure}[t]
    \centering
    \includegraphics[width=\linewidth]{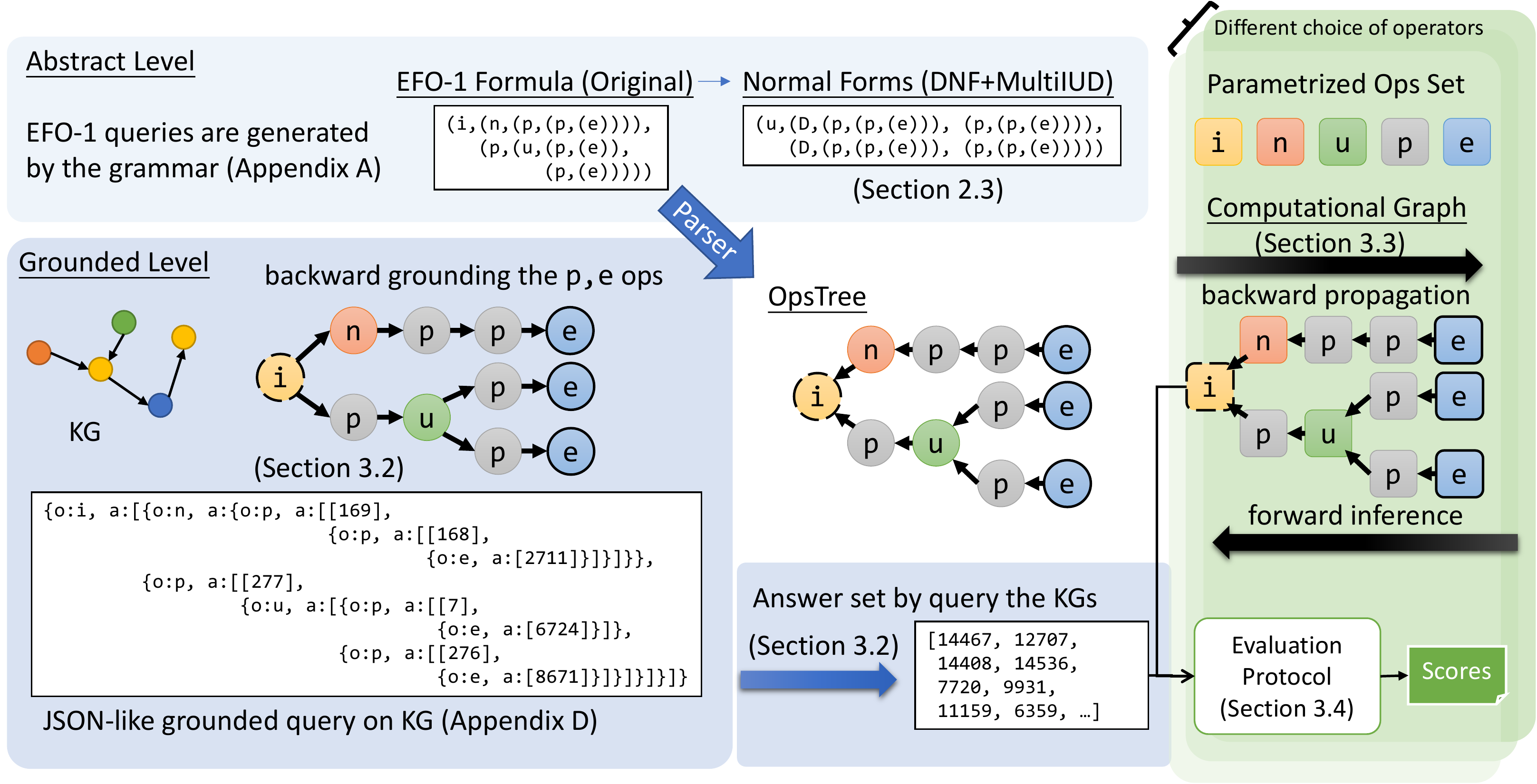}
    \caption{Framework of constructing the EFO-1-QA benchmark. 
    The query types are defined by the EFO-1 formula, which is generated the Grammar 2 in the Appendix~\ref{app:lisp-grammar} and generated at the abstract level. The EFO-1 formula can be converted to different normal forms and represented in the different operators. A parser is employed to produce the OpsTree from the EFO-1 formula. Queries are grounded by the backward DFS of the OpsTree on the full graph and the answers are sampled by the forward execution of the OpsTree on the partial graph as explained in Section~\ref{sec:grounding-sampling}. The OpsTree can also be used to build the computational graph with the parameterized operators, which are used to train and infer the CQA models by the backward propagation and forward inference. Finally, the estimated query embeddings are evaluated by the Evaluation Protocol with 5 metrics given in the Section~\ref{sec:evaluation}.}
    \label{fig:benchmark-pipeline}
\end{figure}

In this section, we cover the detailed framework of the construction of EFO-1-QA benchmark. Our framework includes (1) the generation and normalization of EFO-1 query types following the definitions in Section~\ref{sec:cqa-over-kg}; (2) grounding query types to specific knowledge graph to get the queries and sampling the answer set; (3) constructing the computational graph to conduct the end-to-end training and evaluation. (4) Evaluation the model with metrics that emphasis on the generalizability.
In our practice, we keep the EFO-1-QA dataset as practical as possible and follow the common practice of BetaE dataset. Our benchmark contains 301 different query types (in the Original form) and is at least 20 times larger than the previous works~\cite{DBLP:conf/iclr/RenHL20,DBLP:conf/nips/RenL20,DBLP:conf/kdd/LiuDJZT21}. Moreover, the overall dataset construction and inference pipeline is general enough. It can be applied to EFO-1 queries of any complexity and any properly parametrized operators.  

\subsection{Generation of EFO-1 Query Types}~\label{sec:choice}

Since EFO-1 queries can be represented by the OpsTree, we employ a LISP-like language~\cite{mccarthy1965lisp} to describe the \emph{EFO-1 query types}. The string generated by our grammar is called an \emph{EFO-1 formula}
(see Appendix~\ref{app:lisp-grammar} for more details about the grammars).
Follow our derivation of EFO-1 queries from the FO queries (see Appendix~\ref{app:efo-1}), five operators are naturally introduced by the Skolemization process, including entity \texttt{e} (0 operand), projection \texttt{p} (1 operand), 
negation \texttt{n} (1 operand), intersection \texttt{i} (2 operands), and union \texttt{u} (2 operands). Specifically, Table~\ref{tab:EFO-1-formula-of-beta} give the example of the EFO-1 formulas for 14 query types in the BetaE dataset~\cite{DBLP:conf/nips/RenL20}. We can parse any EFO-1 formula to the OpsTree according the our grammar.

\begin{table}[t]
\centering
\caption{Number of EFO-1 query types with respect to the maximum length of projection chains and number of anchor nodes for EFO-1-QA and BetaE dataset. The boldface indicates the queries that are not discussed sufficiently in the BetaE dataset.}~\label{tab:efo-1-and-beta}
\footnotesize
\begin{tabular}{lrrrrrrrr}
\toprule
\multirow{3}{*}{\shortstack[l]{Max length \\of projection \\ chains}} & \multicolumn{8}{c}{\# anchor nodes}\\ \cmidrule{2-9}
& \multicolumn{4}{c}{EFO-1-QA} & \multicolumn{4}{c}{BetaE} \\\cmidrule{2-9}
        & 1               & 2               & 3             & Sum     & 1     & 2      & 3 & Sum \\\midrule
1       & 1               & 3               & 12             & 16     & 1     & 3      & \textbf{2} & 6\\
2       & 1               & 10               & 91            & 102     & 1     & \textbf{6}     & \textbf{0} & 7\\
3       & 1               & 13              & 169           & 183     & 1    & \textbf{0}    & \textbf{0} & 1\\\midrule
Sum & 3 & 26 & 272 & 301 & 3 & 9 & 2 & 14\\\bottomrule
\end{tabular}
\end{table}

In EFO-1-QA benchmark, the EFO-1 formulas
are generated by a depth first search of the Grammar 2 in the Appendix~\ref{app:lisp-grammar} with the \texttt{[e,p,i,u,n]} operators. The grammar explicitly follows the practice of \emph{bounded negation}. That is, we only generate the negation operator when it is one operand of an intersection operator. The produced OpsTree is binary.

Instead of producing endless query types in the combinatorial spaces of EFO-1 queries, we keep the generated types as realistic as possible by following two practical constraints:
(1), we set the maximum length of projection/negation chains to be 3. That is, we consider no more than three projections/negations in any paths from the target root node to anchor leaf nodes, which follows the \texttt{3p} setting in Table~\ref{tab:EFO-1-formula-of-beta}. (2), we limit the number of anchor nodes to no more than 3, which follows the \texttt{3i} setting in Table~\ref{tab:EFO-1-formula-of-beta}. As a result, we generate 301 different query types, more details can be found in Table~\ref{tab:efo-1-and-beta}.

\subsection{Normalization of EFO-1 Query Types}
Interestingly, in the context of learning based CQA models, the logically equivalent transformation of query types may lead to computationally different structures. 
On the one hand, different choices of operators lead to different parameterizations and generalization performances. For example, the set difference operator~\cite{DBLP:conf/kdd/LiuDJZT21} is reported to perform differently from the negation operator~\cite{DBLP:conf/nips/RenL20}.
On the other hand, different normal forms also affect the learning based CQA models. Specifically, different forms alters the query structure, i.e., OpsTree, and might result in different depths or various number of inputs of the specific operator (see the DNF formula and the DNF+IU formula in the Appendix~\ref{app:normal-forms-example} Table~\ref{tab:normal-forms-example}) and finally affect the performance.
For example, DNF has been claimed to be better than the De Morgan by~\cite{DBLP:conf/nips/RenL20} when evaluating on \texttt{2u} and \texttt{up} queries.

However, the impact of the operators and normal forms are not clearly justified in previous works because they are also entangled with parametrization, optimization, and other issues. Our benchmark, to the best of our knowledge, is the first to justify the impact of operators and normal forms from the aspect of the dataset. Our LISP-like language is general enough to be compatible with all those different query types. Here we list ohow EFO-1-QA benchmark considers the impact from choices of operators (see the Grammar 3 in the Appendix~\ref{app:lisp-grammar}) and normal forms.

\paragraph{(A) Choice of the Operators.} We have introduced the \texttt{[e,p,i,u,n]} operator system by Skolemization. 
In BetaE dataset~\cite{DBLP:conf/nips/RenL20}, multi-intersection operator \texttt{I} and multi-union operator \texttt{U} that accept more than two inputs to conduct the intersection and union are chosen in the \texttt{[e,p,I,U,n]} system. In this case, the ``3i'' type in Table~\ref{tab:EFO-1-formula-of-beta} can be rewritten as \texttt{(I,(p,(e)),(p,(e)),(p,(e)))}.
Moreover, the set difference operator \texttt{d} or the generalized multi-difference operator \texttt{D}
are introduced in~\cite{DBLP:conf/kdd/LiuDJZT21} to replace the negation operator \texttt{n} for EFO-1 queries with the \emph{bounded negation} assumption. The rationale behind the \emph{bounded negation} is that the negation should be bounded by a set intersection operation because the set complement against all entities is not practically useful. So one can replace each intersection-negation structure with the set difference, resulting in \texttt{[e,p,i,u,d]} or \texttt{[e,p,I,U,D]} systems. However, the removal of the negation operator made it impossible to apply the De Morgan's law, which can represent the union operator \texttt{u} by intersection \texttt{i} and negation \texttt{n}. More comment of the operators can also be found in the Appendix~\ref{app:lisp-grammar}. To summarize, we consider 7 choices of operators to represent the EFO-1 queries, see Table~\ref{tab:normal-forms}.

\paragraph{(B) Choice of Normal Forms.}
Normal forms, such as Disjunctive Normal Forms (DNF)~\cite{davey2002introduction}, are equivalent classes of query types. Normalization, i.e., converting queries into normal forms, is effective to reduce the number of query types and rectify the estimation process while preserving the logical equivalence.
The participation of different operator systems makes the choices of normal forms more complicated.
In this work, all 9 different forms with 7 different choices of operators are shown in Table~\ref{tab:normal-forms}. This 9 normal forms are selected by enumerating all possible combinations of operators, see Appendix~\ref{app:reason to choose form}. 
The example of each form and how they are transformed are shown in Table~\ref{tab:normal-forms-example} in the Appendix~\ref{app:normal-forms-example}.
After obtaining a query from the generation procedure, we transform them to DNF and other seven forms. Most of the conversions are straightforward except the conversion from the original form to the DNF.

\begin{table}[t]
\centering
\caption{The normal forms of logical queries, related choice of operators and the number of types of each normal form considered in EFO-1-QA benchmark.}~\label{tab:normal-forms}
\footnotesize
\begin{tabular}{lll}
\toprule
(Normal) Forms  & Operators & Comment \\\midrule
Original      & \texttt{[e,p,i,u,n]}  & Sort multiple operands by the alphabetical order \\
DM     & \texttt{[e,p,i,n]} & Replace the \texttt{u} with \texttt{i,n} by De Morgan's rule  \\
DM + I    & \texttt{[e,p,I,n]} & Replace \texttt{i} in DM by \texttt{I}   \\
Original + d  & \texttt{[e,p,i,u,d]} & Replace \texttt{i-n} structure by binary \texttt{d} operator\\
DNF           & \texttt{[e,p,i,u,n]} & Disjunctive Normal Form derived by the Appendix~\ref{app:forms-transformation}\\
DNF + d    & \texttt{[e,p,i,u,d]} & Replace the \texttt{n} in DNF by binary \texttt{d} \\
DNF + IU & \texttt{[e,p,I,U,n]} & Replace the binary \texttt{i,u} in DNF by \texttt{I,U} \\
DNF + IUd & \texttt{[e,p,I,U,d]} & Replace \texttt{n} in DNF+IU by binary \texttt{d}  \\
DNF + IUD  & \texttt{[e,p,I,U,D]}  & Replace the \texttt{n} in DNF+IU by multi-difference \texttt{D} \\
\bottomrule
\end{tabular}
\vspace{-0.1in}
\end{table}

\subsection{Grounding EFO-1 Queries and Sampling the Answer Sets}\label{sec:grounding-sampling}
Given the specific knowledge graph, we can ground the query types with the containing relations and entities. 
We consider the knowledge graph $\mathcal{G}$ and its training subgraph $\mathcal{G}_{train}$, such that $\mathcal{G}_{train}\subset \mathcal{G}$. To emphasis on the generalizability of CQA models that are trained on $\mathcal{G}_{train}$, the queries are grounded to the entire graph $\mathcal{G}$ and we pick the answers that can be obtained on the $\mathcal{G}$ but not the $\mathcal{G}_{train}$. We note that this procedure follows the protocol in~\cite{DBLP:conf/iclr/RenHL20} and prevents the data leakage.

\paragraph{Grounding Query Types.}
The grounding means to assign specific relations and entities from the $\mathcal{G}$ to the \texttt{p} and \texttt{e} operators in the OpsTree. We conduct the grounding process in the \emph{reverse} order, i.e., from the target root node to the leaf anchor nodes, as shown in Figure~\ref{fig:benchmark-pipeline}.
We first sample an entity as the seed answer at the root node and go through the tree. During the iterating, the inputs of each operator are derived by its output. For the set operators such as intersection, union, and negation, we select the inputs sets while guaranteeing the output. For the projection operator, we sample the relation from the reverse edges in the $\mathcal{G}$ that leaves the specific output entity. For the entity operator, i.e., the anchor nodes, we sample the head entity given the relation and the tail entity. In this way, we ensure grounded queries to have at least one answer. The sampling procedure for the negation operator is a bit more complicated and we leave the details in the Appendix~\ref{app:negation-grounding}.
In order to store the grounded relation and query information, we employ the JSON format to serialize the information. The details of the JSON string can be found in the Appendix~\ref{app:json-serialization}.

\paragraph{Sampling Answer Sets.}
Once the query is grounded, we can sample the answer by the execution of the OpsTree in the full knowledge graph $\mathcal{G}$. The execution procedure of each operator is defined in the Table~\ref{tab:operators}. 
The full answer set $A_{full}$ is obtained on the $\mathcal{G}$ and the trivial answer set $A_{trivial}$ is obtained by sampling the training subgraph $\mathcal{G}_{train}$. As we stated, we focus on the answer set $A = A_{full} - A_{trivial}$ that cannot be obtained by simply memorizing the known training graph $\mathcal{G}_{train}$. 
Specifically, we pick the queries whose answer sizes are between 1 and 100, which follows the practice of BetaE dataset~\cite{DBLP:conf/nips/RenL20}. 

For each query type, we can produce one data sample by a grounding and sampling process. We note that the grounding and sampling process does not rely on a specific graph. In this work, we sample the benchmark dataset on three knowledge graphs, including FB15k~\cite{toutanova2015observed}, FB15k-237~\cite{DBLP:conf/nips/BordesUGWY13}, and NELL~\cite{carlson2010toward} with 5000, 8000, and 4000 queries. More details about how the dataset is organized can be found in the Appeidix~\ref{app:data-format}.

\subsection{From OpsTree to Computational Graph}
Similar to the sampling process where the answers are drawn by the forward computation of the OpsTree, we can also construct the end-to-end computational graph with the parameterized operators in the same topology to estimate the answer embeddings. Therefore, we can train and evaluate the CQA models over the constructed computational graphs of all EFO-1 queries. Practically, we can even use any provided checkpoints to initialize the parameterized operators and conduct the inference. Therefore, the EFO-1-QA provides a general test framework of CQA checkpoints with no need to know how the checkpoints are obtained. 

\subsection{Evaluation Protocol}\label{sec:evaluation}
The CQA models are evaluated by the ranking based metrics in the EFO-1-QA benchmark.
Basically, the ranking of all entities are expected to be obtained after the inference. For example, the entities can be ranked by their ``distances'' to the estimated answer embedding. 
We use following metrics to evaluate the generalizability of CQA models, including MRR and HIT@K that have been widely used in previous works~\cite{DBLP:conf/nips/RenL20,DBLP:conf/iclr/ArakelyanDMC21,DBLP:conf/kdd/LiuDJZT21}.

$\bullet\;${\bf MRR.} For each answer entity in the answer set, we consider its ranking with $\mathcal{E} - A_{full}$. That is, the ranking of the given answer against all non-answer entities. The Mean Reciprocal Rank (MRR) for a query is the average of the MRR of all answers of this query. The MRR of a query can be 1 if all the answers are ranked before the rest non-answer entities. Then the query MRR are averaged to the specific query types or the entire dataset.

$\bullet\;${\bf HIT@K.} Similar to MRR, HIT@K is computed for each answer by its ranking in $\mathcal{E}-A_{full}$ and then averaged for the query. In our practice, we consider $K=1,3,10$.

$\bullet\;${\bf Retrieval Accuracy (RA).} Previous metrics focus on the answer entity against non-answer entities, which deviates from the real-world retrieval task. In this paper, we propose the RA score to evaluate how well a model retrieves the entire answer set. The computation of RA score is decomposed into two steps, i.e., (1) to estimate the size of the answer set as $N$, (2) to compute the accuracy of the top-$N$ answers against the true answer set.

We note that EFO-1-QA also supports the counting task. However, since not all the CQA models are designed to count the number of answers, we assume that the ground-truth of the answer size is known and only consider the second step of computing the RA score in this paper. We call the RA score with the known answer size as the RA-Oracle. Moreover, as this benchmark focuses on the generalization property of CQA models, we do not report the evaluation in the entailment setting~\cite{DBLP:conf/nips/SunAB0C20}.

\begin{table}[t]
\centering
\caption{Review of existing CQA datasets, where * means the DNF/DM is required.}~\label{tab:cqa-dataset}.
\scriptsize
\begin{tabular}{llllllllllllll}
\toprule
\multirow{2}{*}{CQA Dataset} & \multicolumn{9}{c}{Support Operators}                                    & \multirow{2}{*}{\shortstack[l]{Support \\ EPFO}} & \multirow{2}{*}{\shortstack[l]{Support \\ EFO-1}} & \multirow{2}{*}{\shortstack[l]{Num. of \\ Forms}} & \multirow{2}{*}{\shortstack[l]{Num. of  Test \\ Query Types}}  \\\cmidrule{2-10}
                           & \texttt{e}      & \texttt{p}      & \texttt{i}     & \texttt{I}      & \texttt{u}      & \texttt{U}      & \texttt{n}      & \texttt{d}      & \texttt{D}      &                               &  &                              \\\midrule
Q2B dataset~\cite{DBLP:conf/iclr/RenHL20}     & \cmark & \cmark & \cmark & \cmark & \xmark & \xmark & \xmark & \xmark & \xmark & \cmark*                        & \xmark          & 1  & 9       \\
HypE dataset~\cite{DBLP:conf/www/ChoudharyRKSR21}        & \cmark & \cmark & \cmark & \cmark & \xmark & \xmark & \xmark & \xmark & \xmark & \cmark*                        & \xmark    & 1 & 9  \\
BetaE dataset~\cite{DBLP:conf/nips/RenL20}          & \cmark & \cmark & \cmark & \cmark & \xmark & \xmark & \cmark & \xmark & \xmark & \cmark*                        & \cmark*      & 2  & 14  \\
EFO-1-QA (ours)  & \cmark & \cmark & \cmark & \cmark & \cmark & \cmark & \cmark & \cmark & \cmark & \cmark                        & \cmark                 & 9  & 301  \\
\bottomrule
\end{tabular}
\end{table}

\section{Related Datasets and the Comparison to EFO-1-QA Benchmark}

Existing datasets are constructed along with the CQA models,
for the purpose of indicating that their models are capable to solve some certain types of queries by providing a few examples. Thus, those datasets contain very limited query types, normal forms and operators, see Table~\ref{tab:cqa-dataset}. However, EFO-1-QA benchmark focuses on how well CQA models work on the whole EFO-1 query space and considering the  impact of operators and normal forms.

\begin{table}[t]
\centering
\caption{Benchmark results of MRR (\%) on different dataset. The results of the BetaE dataset are obtained from the original paper~\cite{DBLP:conf/nips/RenL20,DBLP:journals/corr/abs-2103-00418}.}
\small
~\label{tab:dataset comparison}
\begin{tabular}{lrrrrrrrrrr}
\toprule
\multirow{2}{*}{CQA   Model} & \multicolumn{1}{c}{\multirow{2}{*}{Dataset}} & \multicolumn{3}{c}{FB15k-237} & \multicolumn{3}{c}{FB15k} & \multicolumn{3}{c}{NELL} \\ \cmidrule{3-11}
 & \multicolumn{1}{c}{} & EPFO & Neg. & ALL & EPFO & Neg. & ALL & EPFO & Neg. & ALL \\ \midrule
\multirow{2}{*}{\shortstack[l]{BetaE\\+DNF+IU}} & BetaE & 20.9 & \textbf{5.4} & 15.4 & 41.6 & \textbf{11.8} & 31.0 & 24.6 & \textbf{5.9} & 17.9 \\
 & EFO-1-QA & \textbf{11.8} & 7.5 & \textbf{9.7} & \textbf{23.7} & 16.8 & \textbf{20.3} & \textbf{12.7} & 8.3 & \textbf{10.6} \\ \midrule
\multirow{2}{*}{\shortstack[l]{LogicE\\+DNF+IU}} & BetaE & 22.3 & \textbf{5.6} & 16.3 & 44.1 & \textbf{12.5} & 32.8 & 28.6 & \textbf{6.2} & 20.6 \\
 & EFO-1-QA & \textbf{12.8} & 8.1 & \textbf{10.5} & \textbf{25.4} & 18.2 & \textbf{21.9} & \textbf{15.6} & 10.4 & \textbf{13.1}
 \\ \bottomrule
\end{tabular}
\end{table}

Table~\ref{tab:efo-1-and-beta} already shows that EFO-1-QA benchmark contains much more query types, supported operators and normal forms than BetaE dataset~\cite{DBLP:conf/nips/RenL20}, thereby provides a \emph{more comprehensive} evaluation result. 
Meanwhile,
we compare results of both BetaE~\cite{DBLP:conf/nips/RenL20} and LogicE~\cite{DBLP:journals/corr/abs-2103-00418} between EFO-1-QA benchmark and BetaE dataset~\cite{DBLP:conf/nips/RenL20} in Table~\ref{tab:dataset comparison}. We note that the EFO-1-QA benchmark is \emph{generally harder} than BetaE dataset when averaging results from all query types on three KGs.
Moreover, our comprehensive benchmark brings us many new insights and helps us to refresh the observations from previous dataset.

{\bf Finding 1: Negation queries ares not significantly harder.}
We further separate the query types into two subgroups, i.e., the EPFO queries and the negation queries.
Table~\ref{tab:dataset comparison} shows that results from two dataset have very different distribution of the scores in those two subgroups. This can be explained by the fact that the five negation query types in the BetaE dataset are biased and cannot represent the general performance of the negation queries.

In short, we can conclude that the EFO-1-QA benchmark is more comprehensive,  generally harder, and fairer than existing datasets.

\section{The Empirical Evaluation of the Benchmark}

In this section, we present the evaluation results of the complex query answering models that are compatible to the EFO-1 queries. 

\subsection{Complex Query Answering Models}

We summarize existing CQA models by their supported operators as well as supported query families in Table~\ref{tab:cqa-model-review}. 
Only three CQA models fully support EFO-1 family by their original implementation. Therefore, in our evaluation, we focus on these models, including BetaE~\cite{DBLP:conf/nips/RenL20}, LogicE~\cite{DBLP:journals/corr/abs-2103-00418}, and NewLook~\cite{DBLP:conf/kdd/LiuDJZT21}. These models are trained on the BetaE training set and evaluated on EFO-1-QA benchmark. Specifically, the BetaE is trained by the original implementation released by the authors~\footnote{\url{https://github.com/snap-stanford/KGReasoning}} and evaluated in our framework. LogicE and NewLook are re-implemented, trained and tested by our framework. The NewLook implementation is adapted to fit into the generalization evaluation, see the Appendix~\ref{app:newlook-implementation}.

\subsection{Benchmark Results}
The benchmark result is shown in Table~\ref{tab:benchmark all} for three models with five supported normal forms in total on three KGs.
Besides the findings in Table~\ref{tab:dataset comparison}, the average HIT@1 of NewLook is reported to be 37.0 in their paper~\cite{DBLP:conf/kdd/LiuDJZT21} but is 10.1 on our EFO-1-QA. This can be caused by the hardness of our dataset and our implementation prevent the data leakage.
We also group the 301 query types into 9 groups by their depth and width. The detailed results of FB15K-237 can be found in Table~\ref{tab:benchmark}. For FB15K and NELL, the corresponding results are listed in the Table~\ref{tab:benchmark-fb15k} and Table~\ref{tab:benchmark-nell} in the Appendix~\ref{app:additional_exp}. Detailed analysis in the Appendix~\ref{app:additional_exp} justifies the impact of query structures, for the first time.

\begin{table}[t]
\centering
\caption{Benchmark results (\%) for all three models and their corresponding normal forms.}
\label{tab:benchmark all}
\small
\begin{tabular}{rrrrrrrrrr}
\toprule
\multirow{2}{*}{\shortstack[l]{Knowledge\\Graph}} & CQA Moddel & \multicolumn{3}{c}{BetaE} & \multicolumn{3}{c}{LogicE} & \multicolumn{2}{c}{NewLook} \\ \cmidrule{3-10}
 & Normal Form & DM & \shortstack[l]{DM\\+I} & \shortstack[l]{DNF\\+IU} & DM & \shortstack[l]{DM\\+I} & \shortstack[l]{DNF\\+IU} & \shortstack[l]{DNF\\+IUd} & \shortstack[l]{DNF\\+IUD} \\ \midrule
\multirow{5}{*}{\shortstack[l]{FB15k\\-237}}  & MRR & 8.48 & 8.50 & {\ul 9.67} & 10.00 & 10.01 & {\ul \textbf{10.46}} & 9.11 & {\ul 9.13} \\
 & HIT@1 & 4.35 & 4.37 & {\ul 4.89} & 5.26 & 5.27 & {\ul \textbf{5.42}} & 4.80 & {\ul 4.81} \\
 & HIT@3 & 8.54 & 8.56 & {\ul 9.69} & 10.19 & 10.21 & {\ul \textbf{10.61}} & 9.14 & {\ul 9.15} \\
 & HIT@10 & 16.25 & 16.27 & {\ul 18.73} & 19.04 & 19.06 & {\ul \textbf{20.01}} & 17.17 & {\ul 17.20} \\
 & RA-Oracle & 11.49 & 11.51 & {\ul 13.69} & 13.63 & 13.65 & {\ul \textbf{14.37}} & 12.43 & {\ul 12.45} \\ \midrule
\multirow{5}{*}{FB15k} & MRR & 17.18 & 17.22 & {\ul 20.31} & 20.53 & 20.55 & {\ul \textbf{21.89}} & 19.80 & {\ul 19.87} \\
 & HIT@1 & 10.46 & 10.51 & {\ul 12.05} & 12.68 & 12.70 & {\ul \textbf{13.14}} & 11.96 & {\ul 11.99} \\
 & HIT@3 & 18.76 & 18.81 & {\ul 22.10} & 22.71 & 22.73 & {\ul \textbf{24.17}} & 21.58 & {\ul 21.66} \\
 & HIT@10 & 30.30 & 30.35 & {\ul 36.74} & 35.93 & 35.96 & {\ul \textbf{39.33}} & 35.28 & {\ul 35.44} \\
 & RA-Oracle & 21.83 & 21.89 & {\ul 27.51} & 26.92 & 26.95 & {\ul \textbf{29.38}} & 26.57 & {\ul 26.66} \\ \midrule
\multirow{5}{*}{NELL} & MRR & 8.93 & 8.94 & {\ul 10.58} & 11.13 & 11.14 & {\ul \textbf{13.07}} & 9.88 & {\ul 9.90} \\
 & HIT@1 & 5.58 & 5.59 & {\ul 6.52} & 7.26 & 7.27 & {\ul \textbf{8.31}} & {\ul 6.04} & {\ul 6.04} \\
 & HIT@3 & 9.38 & 9.39 & {\ul 11.12} & 11.89 & 11.89 & {\ul \textbf{14.01}} & 10.35 & {\ul 10.36} \\
 & HIT@10 & 15.27 & 15.29 & {\ul 18.32} & 18.38 & 18.39 & {\ul \textbf{22.04}} & 17.10 & {\ul 17.13} \\
 & RA-Oracle & 12.08 & 12.09 & {\ul 14.98} & 15.25 & 15.26 & {\ul \textbf{18.39}} & 14.15 & {\ul 14.16} \\ \bottomrule
\end{tabular}
\end{table}

\begin{table}[t]
\caption{Benchmark results(\%) on FB15k-237. The mark $\dagger$ indicates the query groups that previous datasets have not fully covered. The boldface indicates the best scores. The best scores of the same model are underlined.}\label{tab:benchmark}
\scriptsize
\begin{tabular}{lllrrrrrrrrrr}
\toprule
\multirow{2}{*}{\shortstack[l]{CQA\\  Model}} & \multirow{2}{*}{\shortstack[l]{Normal\\ Form}}        & \multirow{2}{*}{Metric} & \multicolumn{9}{c}{Query type groups (\# anchor nodes, max length of Projection chains)}                                                                       & \multirow{2}{*}{AVG.} \\\cmidrule{4-12}
                             &                                     &                         & (1,1)          & (1,2)          & (1,3)          & (2,1)          & (2,2)$^\dagger$          & (2,3)$^\dagger$          & (3,1$^\dagger$)          & (3,2)$^\dagger$          & (3,3)$^\dagger$          &                                    \\\midrule
\multirow{15}{*}{BetaE} & \multirow{5}{*}{DM} & MRR & 18.79 & 9.72 & 9.64 & 12.76 & 8.48 & 8.10 & 11.34 & 8.58 & 8.09 & 8.48 \\
 &  & HIT@1 & 10.63 & 4.63 & 4.68 & 7.07 & 4.13 & 3.89 & 5.99 & 4.42 & 4.16 & 4.35 \\
 &  & HIT@3 & 20.37 & 9.61 & 9.44 & 13.47 & 8.37 & 8.02 & 11.99 & 8.66 & 8.11 & 8.54 \\
 &  & HIT@10 & 36.19 & 19.80 & 19.38 & 24.27 & 16.82 & 16.03 & 21.99 & 16.41 & 15.43 & 16.25 \\
 &  & RA-Oracle & 14.38 & 14.40 & 16.99 & 14.09 & 12.07 & 13.04 & 12.51 & 10.86 & 11.48 & 11.49 \\ \cmidrule{2-13}
 & \multirow{5}{*}{\shortstack[l]{DM\\+I}} & MRR & 18.79 & 9.72 & 9.64 & 12.76 & 8.48 & 8.10 & 11.39 & 8.59 & 8.12 & 8.50 \\
 &  & HIT@1 & 10.63 & 4.63 & 4.68 & 7.07 & 4.13 & 3.89 & 6.05 & 4.43 & 4.19 & 4.37 \\
 &  & HIT@3 & 20.37 & 9.61 & 9.44 & 13.47 & 8.37 & 8.02 & 12.01 & 8.68 & 8.14 & 8.56 \\
 &  & HIT@10 & 36.19 & 19.80 & 19.38 & 24.27 & 16.82 & 16.03 & 22.01 & 16.43 & 15.47 & 16.27 \\
 &  & RA-Oracle & 14.38 & 14.40 & 16.99 & 14.09 & 12.07 & 13.04 & 12.58 & 10.88 & 11.52 & 11.51 \\ \cmidrule{2-13}
 & \multirow{5}{*}{\shortstack[l]{DNF\\+IU}} & MRR & 18.79 & 9.72 & 9.64 & 14.39 & 9.28 & 8.86 & 13.14 & 9.76 & 9.32 & {\ul 9.67} \\
 &  & HIT@1 & 10.63 & 4.63 & 4.68 & 7.78 & 4.48 & 4.20 & 6.83 & 4.93 & 4.72 & {\ul 4.89} \\
 &  & HIT@3 & 20.37 & 9.61 & 9.44 & 15.11 & 9.12 & 8.74 & 13.86 & 9.79 & 9.28 & {\ul 9.69} \\
 &  & HIT@10 & 36.19 & 19.80 & 19.38 & 28.04 & 18.55 & 17.67 & 25.82 & 18.90 & 17.95 & {\ul 18.73} \\
 &  & RA-Oracle & 14.38 & 14.40 & 16.99 & 16.87 & 13.58 & 14.69 & 15.39 & 12.93 & 13.83 & {\ul 13.69} \\ \midrule
\multirow{15}{*}{LogicE} & \multirow{5}{*}{DM} & MRR & 20.71 & 10.70 & 10.18 & 15.66 & 10.01 & 9.41 & 13.71 & 10.12 & 9.54 & 10.00 \\
 &  & HIT@1 & 11.66 & 5.20 & 5.25 & 8.81 & 5.00 & 4.83 & 7.38 & 5.27 & 5.06 & 5.26 \\
 &  & HIT@3 & 23.02 & 10.66 & 9.96 & 16.72 & 10.07 & 9.43 & 14.57 & 10.33 & 9.67 & 10.19 \\
 &  & HIT@10 & 39.81 & 21.25 & 19.48 & 29.66 & 19.66 & 18.12 & 26.33 & 19.38 & 18.04 & 19.04 \\
 &  & RA-Oracle & 15.64 & 15.27 & 17.28 & 17.49 & 13.97 & 14.75 & 15.62 & 12.99 & 13.61 & 13.63 \\ \cmidrule{2-13}
 & \multirow{5}{*}{\shortstack[l]{DM\\+I}} & MRR & 20.71 & 10.70 & 10.18 & 15.66 & 10.01 & 9.41 & 13.76 & 10.14 & 9.56 & 10.01 \\
 &  & HIT@1 & 11.66 & 5.20 & 5.25 & \textbf{8.81} & 5.00 & 4.83 & \textbf{7.41} & 5.28 & 5.07 & 5.27 \\
 &  & HIT@3 & 23.02 & 10.66 & 9.96 & 16.72 & 10.07 & 9.43 & 14.67 & 10.35 & 9.69 & 10.21 \\
 &  & HIT@10 & 39.81 & 21.25 & 19.48 & 29.66 & 19.66 & 18.12 & 26.41 & 19.42 & 18.06 & 19.06 \\
 &  & RA-Oracle & 15.64 & 15.27 & 17.28 & 17.49 & 13.97 & 14.75 & 15.64 & 13.01 & 13.63 & 13.65 \\ \cmidrule{2-13}
 & \multirow{5}{*}{\shortstack[l]{DNF\\+IU}} & MRR & 20.71 & 10.70 & 10.18 & \textbf{15.86} & \textbf{10.27} & \textbf{9.67} & \textbf{14.06} & \textbf{10.56} & \textbf{10.06} & {\ul \textbf{10.46}} \\ 
 &  & HIT@1 & 11.66 & 5.20 & 5.25 & 8.69 & \textbf{5.06} & \textbf{4.87} & 7.36 & \textbf{5.41} & \textbf{5.27} & {\ul \textbf{5.42}} \\
 &  & HIT@3 & 23.02 & 10.66 & 9.96 & \textbf{16.85} & \textbf{10.31} & \textbf{9.66} & \textbf{14.90} & \textbf{10.71} & \textbf{10.16} & {\ul \textbf{10.61}} \\
 &  & HIT@10 & 39.81 & 21.25 & 19.48 & \textbf{30.62} & \textbf{20.26} & \textbf{18.70} & \textbf{27.48} & \textbf{20.39} & \textbf{19.06} & {\ul \textbf{20.01}} \\
 &  & RA-Oracle & 15.64 & 15.27 & 17.28 & \textbf{17.94} & \textbf{14.46} & \textbf{15.31} & \textbf{15.99} & \textbf{13.64} & \textbf{14.48} & {\ul \textbf{14.37}} \\ \midrule
\multirow{10}{*}{NewLook} & \multirow{5}{*}{\shortstack[l]{DNF\\+IUd}} & MRR & \textbf{22.31} & \textbf{11.19} & \textbf{10.39} & 16.02 & 9.46 & 9.29 & 11.54 & 8.62 & 8.95 & 9.11 \\
 &  & HIT@1 & \textbf{13.55} & \textbf{5.62} & \textbf{5.18} & 9.42 & 4.85 & 4.85 & 6.17 & 4.47 & 4.74 & 4.80 \\
 &  & HIT@3 & \textbf{24.62} & \textbf{11.40} & \textbf{10.38} & 17.31 & 9.44 & 9.19 & 12.03 & 8.62 & 8.93 & 9.14 \\
 &  & HIT@10 & \textbf{40.53} & \textbf{22.18} & \textbf{20.47} & 29.10 & 18.20 & 17.58 & 22.21 & 16.34 & 16.76 & 17.17 \\
 &  & RA-Oracle & \textbf{17.66} & \textbf{16.32} & \textbf{17.79} & 17.53 & 13.00 & 14.66 & 12.40 & 10.85 & 12.91 & 12.43 \\ \cmidrule{2-13}
 & \multirow{5}{*}{\shortstack[l]{DNF\\+IUD}} & MRR & \textbf{22.31} & \textbf{11.19} & \textbf{10.39} & 16.02 & 9.46 & 9.29 & 11.59 & 8.65 & 8.96 & {\ul 9.13} \\
 &  & HIT@1 & \textbf{13.55} & \textbf{5.62} & \textbf{5.18} & 9.42 & 4.85 & 4.85 & 6.19 & 4.48 & 4.74 & {\ul 4.81} \\
 &  & HIT@3 & \textbf{24.62} & \textbf{11.40} & \textbf{10.38} & 17.31 & 9.44 & 9.19 & 12.06 & 8.65 & 8.94 & {\ul 9.15} \\
 &  & HIT@10 & \textbf{40.53} & \textbf{22.18} & \textbf{20.47} & 29.10 & 18.20 & 17.58 & 22.33 & 16.41 & 16.78 & {\ul 17.20} \\
 &  & RA-Oracle & \textbf{17.66} & \textbf{16.32} & \textbf{17.79} & 17.53 & 13.00 & 14.66 & 12.43 & 10.88 & 12.92 & {\ul 12.45}    \\\bottomrule                 
\end{tabular}
\end{table}

\section{Analysis of the \texttt{[e,p,i,u,n]} System}
As discussed in Section~\ref{sec:choice}, a CQA model may model queries with multiple choices of operators, which are different in computing while equivalent in logic. We here focus on the canonical choice of \texttt{[e,p,i,u,n]} since this system is naturally derived by Skolemization, represents EFO-1 queries without any assumptions such as \emph{bounded negation}. The best model LogicE in Table~\ref{tab:benchmark all} is picked in this section.

\subsection{Combinatorial Generalizability of Operators}

Since the projection operator plays a pivotal role in query answering as shown in Appendix~\ref{sec:importance of projection}, 
For projection, we train models by \{\texttt{1p}\}, \{\texttt{1p,2p}\}, and \{\texttt{1p,2p,3p}\} queries and evaluate on \texttt{1p,2p,3p,4p,5p}\footnote{\texttt{1p,2p},and \texttt{3p} are shown in Table~\ref{tab:EFO-1-formula-of-beta}, \texttt{4p} and \texttt{5p} are defined similarly.}. The experiment result is shown in the Table~\ref{tab:p_generalize}.
We can see that training on deeper query types benefits the generalization power as the performances on unseen query types are improved. However, the performance on \texttt{1p} decreases at the same time.

For the intersection, we train models by \{\texttt{1p,2i}\} and \{\texttt{1p,2i,3i}\}\footnote{\texttt{1p} is also included in training to ensure the performance the projection.} queries and evaluate on \texttt{2i,3i,4i} queries.\footnote{\texttt{2i} and \texttt{3i} are shown in Table \ref{tab:EFO-1-formula-of-beta}, 4i is defined as \texttt{(i,(i,(i,(p,(e)),(p,(e))),(p,(e))),(p,(e)))}.} As shown in Table \ref{tab:i_generalize}, adding \texttt{3i} to training queries helps with the performance on \texttt{3i,4i} while detriment performance on \texttt{2i}.

{\bf Finding 2: More training query types do not necessarily lead to better performance.}
Adding more queries to train is not helpful to all query types, since it may benefit some query types while impairing others. Our observation indicates the interaction mechanisms between query types is not clear. Thus, how to properly train the CQA models is still open.

{\bf Finding 3: More complex queries do not necessarily have to worse performance.}
We can see that the more complex \texttt{p} queries are, the worse performance they have. However, for \texttt{i} queries, more complex \texttt{i/I} has better performance. In the combinatorial space where those two operators are combined, we cannot even conclude more complex queries have worse performance. This might support our observation that negation queries are not significantly harder since negation operator is assumed to be bounded by an intersection operator.

\begin{table}
\centering
\makebox[0pt][c]{\parbox{1.1\textwidth}{%
    \begin{minipage}[b]{0.46\textwidth}\centering
{
\caption{Generalization performance of projection on FB15k-237 in MRR (\%). 
}
\label{tab:p_generalize}
\centering
\footnotesize
\begin{tabular}{lrrrrr}
\toprule
Training  & 1p & 2p & 3p & 4p & 5p \\ \midrule
1p & 19.36 & 4.98 & 3.95 & 3.17 & 2.93 \\
1p,2p & 19.22 & 9.01 & 7.98 & 7.22 & 7.15 \\
1p,2p,3p & 17.81 & 9.45 & 9.59 & 9.52 & 9.32 \\ \bottomrule
\end{tabular}%
}
    \end{minipage}
    \hspace{0.1in}
    \begin{minipage}[b]{0.56\textwidth}\centering
{
\caption{Generalization performance of intersection on FB15k-237 in MRR (\%).}
\label{tab:i_generalize}
\centering
\footnotesize
\begin{tabular}{lrrrrrr}
\toprule
\multirow{2}{*}{\shortstack[l]{Training }} & \multicolumn{3}{c}{multi-input \texttt{I}} & \multicolumn{3}{c}{binary input \texttt{i}}\\\cmidrule{2-7}
& 2i & 3i & 4i  & 2i & 3i & 4i  \\ \midrule
1p,2i & 32.24 & 41.66 & 52.37  & 32.24 & 41.66 & 51.78  \\ 
1p,2i,3i & 31.97 & 42.67 & 52.70 & 31.97 & 42.32 & 52.10 \\ \bottomrule
\end{tabular}%
}
    \end{minipage}%
}}
\end{table}



\subsection{Impact of the Normal Forms}
To study the impact of different normal forms, except for the averaged results in Table\ref{tab:benchmark all}, we also compares every normal forms with LogicE~\cite{DBLP:journals/corr/abs-2103-00418} as our evaluation model and the results are shown in Table~\ref{tab:normal form impact} and Table~\ref{tab:normal-form-num} in the Appendix~\ref{app:normal-form-num}.

$\bullet\;${\bf DM vs. DNF.}
Formulas with unions can be modeled in two different ways: (1) transformed into Disjunctive Normal Form (DNF) as showed in Appendix~\ref{app:forms-transformation}, (2) with union converted to intersection and negation by the De Morgan's law (DM). In Table~\ref{tab:normal form impact}, we find that DNF outperforms DM in about 90\% cases, whether DM uses \texttt{I} or not. However, there are still some cases where DM can outperform DNF.


$\bullet\;${\bf Original vs. DNF+IU.} DNF+IU outperforms all other normal forms. Moreover, it is a universal form to support all circumstances, making itself the most favorable form. Interestingly, the original form, meanwhile, outperform the DNF, suggesting it has its own advantage in modeling.

{\bf Finding 4: There is no rule of thumb for choosing the best normal form.} When evaluated on BetaE dataset, one may observe that the DNF is always better than DM. However, in EFO-1-QA, our evaluation shows that there is no normal form that can outperform others in every query types. Thus, how to choose the normal form for specific query type to obtain the best inference-time performance is also an open problem.

\begin{table}[t]
\small
\centering
\caption{Impact of normal forms of LogicE on FB15k-237. Each cell indicates the winning rate of the form by its row against the form by its column.}
\label{tab:normal form impact}
\footnotesize
\begin{tabular}{lrrrrr}
\toprule
Outperform   Rate \% & Original & DM & DM+I & DNF & DNF+IU \\ \midrule
Original & 0.00 & 85.96 & 60.61 & 53.33 & 43.33 \\
DM & 14.04 & 0.00 & 41.33 & 12.23 & 20.31 \\
DM+I & 39.39 & 58.67 & 0.00 & 28.50 & 11.60 \\
DNF & 46.67 & 87.77 & 71.50 & 0.00 & 41.67 \\
DNF+IU & 56.67 & 79.69 & 88.40 & 58.33 & 0.00 \\ \bottomrule
\end{tabular}
\end{table}

\section{Conclusion}

In this paper, we present a framework to investigate the combinatorial generalizability of CQA models. With this framework, the EFO-1-QA benchmark dataset is constructed. Comparisons between existing dataset shows that EFO-1-QA data is more comprehensive, generally harder and fairer. The detailed analysis justifies, for the first time, the impact of the choices of different operators and normal forms. Notably, our evaluation leads four insightful findings that refreshes the observations on previous datasets. Two findings also leads to the open problems for training and inference the CQA models. We hope that our framework, dataset, and findings can facilitate the related research towards combinatorial generalizable CQA models.

\section*{Acknowledgement}
The authors of this paper were supported by the NSFC Fund (U20B2053) from the NSFC of China, the RIF (R6020-19 and R6021-20) and the GRF (16211520) from RGC of Hong Kong, the MHKJFS (MHP/001/19) from ITC of Hong Kong.

\bibliographystyle{plainnat}

\newpage
\appendix

\begin{center}
    \Huge\bf
    Appendix
\end{center}

\section{The Formal Definition and Derivation of EFO-1 Query Family}~\label{app:efo-1}

\subsection{First Order Queries, A Self-Contained Guide}

The First-Order (FO) query is a very expressive family of logical queries given by the following definitions. 

The first-order logic handles the set of variables $Vars = \{x_1, ..., x_n\}$ and set of functions $\{f_1, ..., f_m\}$. We say $f$ is a function of arity $k\geq 0$ if $f$ has $k$ inputs. Functions of arity 0 are called constants. Then we give the formal definition of terms.
\begin{definition}[$Terms$]
The set of \textbf{terms} is defined inductively as follows:
\begin{itemize}
    \item $Vars \subset Terms$.
    \item If $t_1, ..., t_k \in Terms$ and $f$ is a $k$-ary function, then $f(t_1,...,t_k)\in Terms$.
    \item Nothing else in $Terms$.
\end{itemize}
\end{definition}

In first-order logic we also have the predicates $\{P_1, ..., P_l\}$. The predicate $P$ is of arity $k$ when it takes $k$ terms as the input. Each predicate indicates whether the specific type of relation exists amongst its inputs by returning True of False. We note that a predicate of arity 0 is a proposition of the propositional logic. Then we give the formal definition of the first-order formula.

\begin{definition}[First-Order Formula]
The set of \textbf{Formulas} can be defined inductively as follows:
\begin{itemize}
    \item If $t_1, ..., t_k\in Terms$ and $P$ is a $k$-ary predicate, then $P(t_1, ..., t_k) \in Formulas$. (Atomic formulas).
    \item If $\phi\in Formulas$ and $\psi \in Formulas$, then
        \begin{itemize}
            \item $\lnot \phi \in Formulas$,
            \item $\phi \land \psi \in Formulas$,
            \item $\phi \lor \psi \in Formulas$,
        \end{itemize}
        where $\land, \lor$, and $\lnot$ are connectives. We note that in some definition, one may consider the logical implication connective $\implies$. However, our definition is complete since implication can be represented by $\land, \lor$, and $\lnot$.
    \item If $\phi \in Formulas$ and $x\in Vars$, then
        \begin{itemize}
            \item $\exists x.\phi \in Formulas$,
            \item $\forall x.\psi \in Formulas$,
        \end{itemize}
        where $\forall$ and $\exists$ are the universal and the existential quantifiers, respectively.
\end{itemize}
\end{definition}

A first order formula can be converted to various normal forms. The key idea of normal form is that the derived formula is logically equivalent to the original formula. In formally, the prenex normal form is to move all the quantifiers before in the front. Here we give the formal definition of the prenex normal form.

\begin{definition}[Prenex Normal Form]
The prenex normal form of a formula is derived by executing the following six conversions until there is nothing to do.
\begin{itemize}
    \item Convert $\lnot (\forall x.\phi) $ to $\exists x. (\lnot \phi) $;
    \item Convert $\lnot (\exists x.\phi) $ to $\forall x. (\lnot \phi) $;
    \item Convert $(\forall x.\phi) \land \psi $ to $\forall x.(\phi \land \psi) $;
    \item Convert $(\forall x.\phi) \lor \psi $ to $\forall x.(\phi \lor \psi) $;
    \item Convert $(\exists x.\phi) \land \psi $ to $\exists x.(\phi \land \psi) $;
    \item Convert $(\exists x.\phi) \lor \psi $ to $\exists x.(\phi \lor \psi) $;
\end{itemize}
where $\phi, \psi\in Formula$.
\end{definition}

Then we can give rise to the formal definition of the first-order query over the knowledge graph by considering the following first-order formula in the prenex normal form form~\cite{robinson2001handbook}:
\begin{align}
F = \square y_1,...,\square y_m.f(x_1, ..., x_n, y_1, ..., y_m; P, O),
\end{align}
where $\square$ is either the existential quantifier $\exists$ or the universal quantifier $\forall$, $f$ is the first-order formula with logical connectives ($\land,\lor,\lnot$), predicates $p\in P$, and constant object $o\in O$. $\{x_1, ..., x_n\}$ are $n>0$ free logical variables and $\{y_1, ... y_m\}$ are $m$ quantified logical variables.
In the knowledge graph, the predicate $p\in P$ is related to the specific relation $r$. $p(a, b)$ is True if and only if the entity $a, b$ has the relation $r$.

The answer set $\mathcal{A}_F$ of $F$ contains $n$-tuples of objects $A = (a_1, ..., a_n)\in \mathcal{A}_F$, such that, any instantiation of $F$ by considering the assignment of free variables $x_i = a_i, 1\leq i \leq n$ is true if and only if $(a_1, ..., a_n)\in \mathcal{A}_F$.

\subsection{EFO-1 Queries on KG}
We consider a knowledge graph $\mathcal{G} = (\mathcal{E}, \mathcal{R})$, where $\mathcal{E}$ is the set of entities and $\mathcal{R}$ is the set of relation triples. Then we formally list the conditions that narrow FO queries down to Existential First Order Query with the single free variable (EFO-1 queries). Our conditions follow what have been considered in~\cite{DBLP:conf/nips/RenL20,DBLP:journals/corr/abs-2103-00418,DBLP:conf/kdd/LiuDJZT21,DBLP:conf/nips/HamiltonBZJL18},\footnote{This family is also called by Skolem set logic in~\cite{DBLP:journals/corr/abs-2103-00418}.} including:

\begin{compactenum}
\item[(1)] only existential quantifiers $\square = \exists$,
\item[(2)] predicates induced by relations: $p_{\text{rel}}(a_{\text{head}}, a_{\text{tail}}) = \text{True} \iff (\text{head}, \text{rel}, \text{tail})\in \mathcal{R}$, and
\item[(3)] single free logical variable in each query.
\item[(4)] Exists a feasible topological ordering determined by predicate is $O<y_1<\dots<y_m<x$, ($p(a,b)\in F$ gives a partial ordering of $a<b$), and we require that for $\forall y_i, i=1\dots m$,  at least one of the followings is true:

    \begin{itemize}
        \item $\exists p \in P, \exists t, 1\leq t<i, \text{ s.t. } p(y_t, y_i) \in F$
        \item $\exists p\in P, o\in O, \text{ s.t. } p(o,y_i)\in F $
    \end{itemize}
    
    Similarly, at least one of the following must be true:
        \begin{itemize}
        \item $\exists p \in P, \exists j, i < j \leq m , \text{ s.t. } p(y_i,y_j) \in F$
        \item $\exists p\in P, \text{ s.t. } p(y_i,x)\in F $
    \end{itemize}

\end{compactenum}

Based on (1), one can conduct Skolemization~\cite{robinson2001handbook} by replacing all existentially quantified logical variables by corresponding Skolem functions.
The intuition of Skolemization is to replacement of $\exists y$ by a concrete choice function computing $y$ from all the arguments $y$ depends on.
The chosen function is also known as the Skolem function.
Specifically, in the context of knowledge graph a Skolem function is induced from the predicate $p$ by producing all entities that satisfy $p$ given the known entity. In this paper, we also use the term ``projection'' to indicate the Skolem function following~\cite{DBLP:conf/iclr/RenHL20,DBLP:conf/nips/RenL20}.

By (2), all Skolem functions are equivalent to the forward and backward relations in $\mathcal{G}$. According to (3), there will be only one target node. 
Then by (4), guaranteed by the topological ordering and following requirements, we will obtain a tree of operators (OpsTree) , whose root represents the target variable and leaves are the known entities, i.e., anchor nodes. An example of EFO-1 query is shown in Figure~\ref{fig:example}.

Though some work~\cite{DBLP:conf/nips/RenL20} claimed to consider the ``first-order query'' and represent them in the OpsTree, they actually considers the EFO-1 queries that formally derived in this paper. Our formal derivation of EFO-1 queries from the first-order queries shows that there are still gaps towards the truely first order queries. Thus, we still have a long way to achieve the logical completeness.

\section{The LISP-like Grammar for EFO-1 Query Types}~\label{app:lisp-grammar}

Here we present the LISP-like grammar for the EFO-1 query types. The string generated by our grammar is called an EFO-1 formula. Each EFO-1 formula is a set of nested arguments segmented by parentheses. The first argument indicates the specific operator and the other arguments (if any) are the operands of the specific operator. Our grammar is context free and the preliminary version considers the \texttt{[e,p,i,u,n]} operators.
\begin{verbatim}
Grammar 1: EFO-1
    Formula      := Intersection|Union|Projection|Negation
    Intersection := (i,Formula,Formula)
    Union        := (u,Formula,Formula)
    Negation     := (n,Formula)
    Projection   := (p,Formula|Entity)
    Entity       := (e)
\end{verbatim}
where | is the pipe symbol to indicate multiple available replacement.

In our implementation, we follow the assumption of \emph{bounded negation}, where the \texttt{Negation} only appears in one of the operands of intersection. Then, the grammar is modified into
\begin{verbatim}
Grammar 2: EFO-1 with the bounded negation
    Formula      := Intersection|Union|Projection
    Intersection := (i,Formula,Formula|Negation)
    Union        := (u,Formula,Formula)
    Negation     := (n,Formula)
    Projection   := (p,Formula|Entity)
    Entity       := (e)
\end{verbatim}

Moreover, the LISP-like grammar is flexible enough and can be extended by other operators. For example, the following grammar supports multiIUD and bounded negation.
\begin{verbatim}
Grammar 3: Extended EFO-1 with the bounded negation
    Formula            := Intersection|Union|Projection|Difference
                          |Multi-Intersection|Multi-Union|Multi-Difference

    Difference         := (d,Formula,Formula)
    Intersection       := (i,Formula,Formula)
    Union              := (u,Formula,Formula)
    Negation           := (n,Formula)
    Projection         := (p,Formula|Entity)
    Entity             := (e)

    Multi-Intersection := (I,Multi-I-Operands)
    Multi-Union        := (U,Multi-U-Operands)
    Multi-Difference   := (D,Multi-D-Operands)
    Multi-I-Operands   := (Formula|Negation),Multi-I-Operands
                          |(Formula|Negation),Formula
    Multi-U-Operands   := Formula,Multi-U-Operands
                          |Formula,Formula
    Multi-D-Operands   := Formula,Multi-D-Operands
                          |Formula,Formula
\end{verbatim}

Based on the formal description above, we are able to discuss the combinatorial space of EFO-1 queries as well as various normal forms with different choice of operators.

\subsection{Justification of Operators}
Here we present the precise definitions of the operators that are considered in our grammars in the Table~\ref{tab:operators}. The operators can be directly used in the sampling process in Section~\ref{sec:grounding-sampling}. The operator \texttt{e} is also known as the anchor node in this paper.

\subsection{Generate EFO-1 Formulas}~\label{app:generation-remark}
We employ the depth-first search based algorithm to iterate through the Grammar 2. The generated formula can be considered as a binary tree. The generation process terminates when the number of Projection \texttt{p} and Negation \texttt{n} reaches the threshold. The termination threshold is not consistent to the grouping criteria in the Table~\ref{tab:efo-1-and-beta} which only considers the Projection operator. For the generation, we consider the negation additionally in order to avoid the endless generation of negations.

\begin{table}[t]
\caption{The operators that are considered in the grammar of EFO-1 formula.}\label{tab:operators}
\begin{tabular}{lp{3.5cm}lp{3cm}}
\toprule
Operator & Input                                         & Output                            & Explanation                                          \\\midrule
\texttt{e}        & Entity id $e_i$                               & Set $S = \{e_i\}$                   & Create a singleton set from one entity               \\
\texttt{p}        & Entity set $S$ and projection function $Proj$ & Set $T = \{Proj(e): e \in S\}$      & Project a set of entities to another set of entities \\
\texttt{i}        & Entity sets $S_1$ and $S_2$                   & Set $T = S_1 \cap S_2$           & Take the intersection of two sets                    \\
\texttt{u}        & Entity sets $S_1$ and $S_2$                   & Set $T = S_1 \cup S_2$           & Take the union of two sets                           \\
\texttt{n}        & Entity set $S$                                & Set $T = \bar{S}$                 & Take the complement of one set against all entities  \\
\texttt{d}        & Entity sets $S_1$ and $S_2$                   & Set $T = S_1 - S_2$                & Take the set difference of $S_1$ and $S_2$           \\
\texttt{I}        & Entity sets $S_1, S_2,...,S_n$                & Set $T= \cap_{k=1}^n S_k$         & Take the intersection of $n$ sets. $n > 2$           \\
\texttt{U}        & Entity sets $S_1, S_2,...,S_n$                & Set $T= \cup_{k=1}^n S_k$         & Take the union of $n$ sets. $n > 2$                  \\
\texttt{D}        & Entity sets $S_1, S_2,...,S_n$                & Set $T= S_1 - S_2 - \cdots - S_n$ & Take the difference of multiple sets                \\\bottomrule
\end{tabular}
\end{table}

\section{Transformation to Disjunctive Normal Form}~\label{app:forms-transformation}

Transforming the EFO-1 queries to DNF is more complicated than EPFO queries considered in~\cite{DBLP:conf/iclr/RenHL20}, even with the straight forward \texttt{[e,p,i,u,n]} operators. Notably, the order of some operators must not be changed for EFO-1 queries, such as \texttt{i} \& \texttt{p}\footnote{$f(A\cap B) \subset f(A)\cap f(B)$, where $f$ is the projection function and $A$ and $B$ are sets.} or \texttt{n} \& \texttt{p}\footnote{$\overline{f(A)} \subset f(\overline{A})$ where $f$ is the projection function, $\overline{\cdot}$ is the negation operator and $A$ is a set.}. Here we list the procedures to transform general EFO-1 queries with \texttt{[e,p,i,u,n]} to a DNF.
\begin{compactdesc}
\item[1. De Morgan's Law] If the operand of a negation operator is intersection or union, switch the positions of negation and intersection/union with De Morgan's law.
\item[2. Negation cancellation] If the operand of a negation operator is another negation, remove those two negation operators.
\item[3. \texttt{p-u} switch] If the operand of a projection operator is an union, switch the position of union and projection. The projection operator should be duplicated while switching.
\item[4. \texttt{i-u} switch] If one of the operands of an intersection operator is an union, We apply $A\cap(B\cup C) = (A\cap B)\cup (A\cup C)$ to switch the union operator and the intersection operator.
\end{compactdesc}
The first two procedures make the negation operator lower than union and intersection operator and the last two procedures make the union higher than the intersection operator. We follow those 4 procedures until no more changes happens. In this way, we get a DNF of the original formula. 
Since some operators cannot be switched, the DNF of EFO-1 query types may not ensure all intersection operators right below the unions. But for the DNF of first order logical formulas, all conjunctions are right below the disjunctions, see Example~\ref{eg:dnf-of-efo1}. We note that this type of queries has not been discussed so far in the current literature.
\begin{example}\label{eg:dnf-of-efo1}
Considering the DNF of EFO-1 queries
\begin{align}
f_1(f_2(A) \cap f_3(B)) \cup f_4(C)
\end{align}
where $f_i$ is the projection functions and $ABC$ are sets. The query type can be presented by \texttt{(u,(p,(i,(p,(e)),(p,(e)))),(p,(e))}. We note that
This EFO-1 query can be re-written as the first order formula with the single free variable $\alpha$ and the existential quantified variable $\beta$.
\begin{align}
    ? \alpha \exists \beta [p_1(\beta, \alpha) \land p_2(A, \beta)\land p_3(B,\beta )]\lor p_4(C, \alpha),
\end{align}
where $p_i$ are the corresponding predicates from projection function $f_i$, $i=1,...,5$. Though the EFO-1 query has intersection as the input of $f_1$, we still call this query a DNF.
\end{example}
We note that the DNF transformation will exponentially increase the complexity of the queries because of the step 4~\cite{DBLP:conf/nips/RenL20}.

\section{Example of Different Normal Forms}\label{app:normal-forms-example}
An example is also presented to show the difference of different normal forms in the Table~\ref{tab:normal-forms-example}. We note that the example contains 5 anchor nodes to reveal all the features though in EFO-1-QA dataset, sampled data only contains no more than 3 anchor nodes.

\begin{table}[h]
\scriptsize
\centering
\caption{An example for different normal forms, operator systems and an example. }~\label{tab:normal-forms-example}
\begin{tabular}{lll}
\toprule
Normal Forms  & Operator System & EFO-1 Formula Example (indented for clearance)\\\midrule
Original      & \texttt{[e,p,i,u,n]}       & \begin{minipage}{5cm}\begin{verbatim}
(i,(i,(n,(p,(e))),
      (p,(i,(n,(p,(e))),
            (p,(e))))),
   (u,(p,(e)),
      (p,(p,(e)))))
\end{verbatim}\end{minipage}\\
DM      & \texttt{[e,p,i,n]}         &
\begin{minipage}{5cm}\begin{verbatim}
(i,(i,(n,(p,(e))),
      (p,(i,(n,(p,(e))),
            (p,(e))))),
   (n,(i,(n,(p,(e))),
         (n,(p,(p,(e)))))))
\end{verbatim}\end{minipage}\\
DM+I & \texttt{[e,p,I,n]}         &
\begin{minipage}{5cm}\begin{verbatim}
(I,(n,(p,(e))),
   (p,(i,(n,(p,(e))),
         (p,(e)))),
   (n,(i,(n,(p,(e))),
         (n,(p,(p,(e)))))))
\end{verbatim}\end{minipage}\\
DNF          & \texttt{[e,p,i,u,n]}       & \begin{minipage}{5cm}\begin{verbatim}
(u,(i,(i,(i,(n,(p,(p,(e)))),
            (p,(p,(e)))),
         (n,(p,(e)))),
      (p,(e))),
   (i,(i,(i,(n,(p,(p,(e)))),
            (p,(p,(e)))),
         (n,(p,(e)))),
      (p,(p,(e)))))
\end{verbatim}\end{minipage} \\
Original + d   & \texttt{[e,p,i,u,d]}       &
\begin{minipage}{5cm}\begin{verbatim}
(i,(d,(p,(d,(p,(e)),
            (p,(e)))),
      (p,(e))),
   (u,(p,(e)),
      (p,(p,(e)))))
\end{verbatim}\end{minipage} \\
DNF + d    & \texttt{[e,p,i,u,d]}       &
\begin{minipage}{5cm}\begin{verbatim}
(u,(i,(d,(d,(p,(p,(e))),
            (p,(p,(e)))),
         (p,(e))),
      (p,(e))),
   (i,(d,(d,(p,(p,(e))),
            (p,(p,(e)))),
         (p,(e))),
      (p,(p,(e)))))
\end{verbatim}\end{minipage}\\
DNF + IU & \texttt{[e,p,I,U,n]}       & 
\begin{minipage}{5cm}\begin{verbatim}
(U,(I,(n,(p,(e))),
      (n,(p,(p,(e)))),
      (p,(e)),
      (p,(p,(e)))),
   (I,(n,(p,(e))),
      (n,(p,(p,(e)))),
      (p,(p,(e))),
      (p,(p,(e)))))
\end{verbatim}\end{minipage}\\
DNF+IUd  & \texttt{[e,p,I,U,d]}       & \begin{minipage}{5cm}\begin{verbatim}
(U,(d,(d,(I,(p,(e)),
            (p,(p,(e)))),
         (p,(e)))
      (p,(p,(e)))),
   (d,(d,(I,(p,(p,(e))),
            (p,(p,(e)))),
         (p,(e)))
      (p,(p,(e)))))
\end{verbatim}\end{minipage}\\   
DNF+IUD  & \texttt{[e,p,I,U,D]}       & \begin{minipage}{5cm}\begin{verbatim}
(U,(D,(I,(p,(e)),
         (p,(p,(e)))),
      (p,(e)),
      (p,(p,(e)))),
   (D,(I,(p,(p,(e))),
         (p,(p,(e)))),
      (p,(e)),
      (p,(p,(e)))))
\end{verbatim}\end{minipage}\\\bottomrule                 
\end{tabular}
\end{table}

\newpage

\section{JSON Serialization of Grounded Queries}\label{app:json-serialization}

Compared to LISP-like description of the query types, the serialization of the grounded queries should also keep the grounded relations and entities. We employ the JSON format to store the query structure and the instantiation. For each query, we maintain two key value pairs. The first key is \texttt{o}, which indicates the operator and has the string object from \texttt{e,p,i,u,n,d,I,U,D}. The second key is \texttt{a}, which indicates the arguments as a list object. For the grounded projection operator, the first argument is the corresponding relation id and the second argument is another JSON string of its input. For the grounded entity operator, the only argument is the corresponding entity id. For other logical or set operators, their arguments are the strings for the inputs in the JSON format.

\section{Grounding Query Types with Meaningful Negation}~\label{app:negation-grounding}

When grounding the query types, we can barely require the query to be valid:
for example, a query like 'Find one that has won the Oscar but not the Turing award.' is valid while the 'but not the Turing award' part is meaningless since there's actually no one who wins both the Oscar and the Turing award. Therefore, a better alternative should be 'Find one who wins the Oscar but not the Golden Globe.' in consideration of this reason. With the \emph{bounded negation} assumption, and for the sake of simplicity, we may only consider the case of (i, (subquery1), (n, subquery2)) to illustrate our sampling method: we need to finish the sampling in the subqeuery1 first, then we randomly select an entity in the answer set of subquery1 and require the subquery2 to exclude it if possible.

This sampling method creates more realistic grounded queries and those grounded queries are apparently harder in our experiments since those negation queries must be understood by the model to get the correct answer set.

\section{Dataset Format}~\label{app:data-format}

We organize our dataset conceptually into two tables. The first table stores the information about query types, the columns of which include different normal forms of the formula in LISP-language, the formula ID, and other statistics. The second table is to store the information of the grounded queries with columns for easy and hard answer set, JSON string for different normal forms and the formula ID that indicating the query type. Those two tables can be joined by the formula ID. In our practice, we split the second table by the formula ID and store each sub-table in a file named by the formula ID. The data sample is also provided with the supplementary materials.

\section{The Choice of Normal Forms}~\label{app:reason to choose form}

Here we explain the reasons why we choose those normal forms and in Table~\ref{tab:normal-forms}.
The key difference is the way we model the union operator: DM form replace union with intersection and negation while DNF form requires union operator only appear at the root of the OpsTree, which create two basic types of forms.
The original form is also kept for the purpose of comparison.
Then we face the choice of \texttt{IU/iu} and the choice of \texttt{d/D/n}, which can make six combinations in total.

In DNF forms, the combination \texttt{[e,p,i,u,D]} are filtered out for it lacks practical meaning. More importantly, it is either the same as \texttt{[e,p,I,U,D]} or the same as \texttt{[e,p,i,u,d]} when the number of anchor node  is restricted to be no more than three. 

In DM forms, the difference is not allowed as it violates the hypothesis of \emph{bounded negation}. The choice of \texttt{I/i} offers us two variants: DM and DM+\texttt{I}.

In original forms, we only offer the choice of \texttt{d/n} to avoid complex transformation and keep the queries as `original' as possible.

In total: five DNF forms, two DM forms, and two original forms makes  nine forms we listed in Table~\ref{tab:normal-forms}.

\section{Implementation Details of NewLook}\label{app:newlook-implementation}
Some details of NewLook is not fully covered in the original paper~\cite{DBLP:conf/kdd/LiuDJZT21} while the released version\footnote{\url{https://github.com/lihuiliullh/NewLook}} is not suitable to justify the combinatorial generalizability. Here we list several details of our implementation of NewLook.
\begin{compactdesc}
\item[Group Division] In the released version, the connectivity tensor $\tau$ is generated by entire graph. Accessing the entire graph contradicts the Open Word Assumption. In our reproduction, we only use the training edges to avoid leaking the information of unseen edges and the number of group is set to 200. 
\item[Intersection] The origin mathematical formulas have been proven to be impractical since $z_i$ will be infinity when $x_{u_{i}}=x_{u_{t}}$, and we fix this problem by adding a small value $\epsilon=0.01$ in the denominator.
\item[Difference] The update method of x has not been shown, so we implement this by intuition: $(x_{u_{t}})_{j}=1 \iff (x_{u_{1}})_{j}=1, \quad(x_{u_{i}})_{j}=0 ,\quad \forall i =2,\dots, k$.
\item[Loss function] The value of hyper-parameter $\lambda$ is not provided, so we set it to 0.02.
\item[MLP] The hyperparameter of MLP is not given, so we let all MLP be a two-layer network with hidden dimension as 1,600 and ReLU activation.
\end{compactdesc}



\section{Short Review of Existing CQA Models}

\begin{table}[t]
\centering
\caption{Review of existing CQA models, where * means the DNF is required and $\dagger$ means the bounded negation must be assumed. For EmQL, the difference operator is claimed to be available, however, it does not indicate how to train this operator as intersection and union.}~\label{tab:cqa-model-review}
\footnotesize
\begin{tabular}{llllllllllll}
\toprule
\multirow{2}{*}{CQA Model} & \multicolumn{9}{c}{Support Operators}                                          & \multirow{2}{*}{\shortstack[l]{Support \\ EPFO}} & \multirow{2}{*}{\shortstack[l]{Support \\ EFO-1}} \\\cmidrule{2-10}
                           & \texttt{e}      & \texttt{p}      & \texttt{i}     & \texttt{I}      & \texttt{u}      & \texttt{U}      & \texttt{n}      & \texttt{d}      & \texttt{D}      &                               &                                \\\midrule
GQE~\cite{DBLP:conf/nips/HamiltonBZJL18}    & \cmark & \cmark & \cmark & \cmark & \xmark & \xmark & \xmark & \xmark & \xmark & \cmark*                        & \xmark                         \\
Q2B~\cite{DBLP:conf/iclr/RenHL20}           & \cmark & \cmark & \cmark & \cmark & \xmark & \xmark & \xmark & \xmark & \xmark & \cmark*                        & \xmark                         \\
EmQL~\cite{DBLP:conf/nips/SunAB0C20}        & \cmark & \cmark & \cmark & \xmark & \cmark & \xmark & \xmark & $\square$ & \xmark & \cmark                        & $\square^\dagger$                         \\
BiQE~\cite{DBLP:conf/aaai/KotnisLN21}    & \cmark & \cmark & \cmark & \cmark & \xmark & \xmark & \xmark & \xmark & \xmark & \cmark*                        & \xmark                         \\
HypE~\cite{DBLP:conf/www/ChoudharyRKSR21}        & \cmark & \cmark & \cmark & \cmark & \xmark & \xmark & \xmark & \xmark & \xmark & \cmark*                        & \xmark                   \\
BetaE~\cite{DBLP:conf/nips/RenL20}          & \cmark & \cmark & \cmark & \cmark & \xmark & \xmark & \cmark & \xmark & \xmark & \cmark*                        & \cmark*                         \\
CQD~\cite{DBLP:conf/iclr/ArakelyanDMC21}    & \cmark & \cmark & \cmark & \xmark & \cmark & \xmark & \xmark & \xmark & \xmark & \cmark                       & \xmark                         \\
LogicE~\cite{DBLP:journals/corr/abs-2103-00418} & \cmark & \cmark & \cmark & \cmark & \xmark & \xmark & \cmark & \xmark & \xmark & \cmark*                        & \cmark*                         \\
NewLook~\cite{DBLP:conf/kdd/LiuDJZT21} & \cmark & \cmark & \cmark & \cmark & \xmark & \xmark & \xmark & \cmark & \cmark & \cmark* & \cmark*$^\dagger$\\\bottomrule                        
\end{tabular}
\end{table}

We compared the details of different CQA models in Table~\ref{tab:cqa-model-review}. We can see that most of the CQA models support the EPFO queries but only three CQA models support EFO-1 query.

\section{The Normal Forms in \texttt{[e,p,i,u,n]} System}\label{app:normal-form-num}
To justify the impact of normal forms, we select the query types that has different EFO-1 formula representation when two normal forms are fixed. The number of the query types that are picked given two normal forms are shown in the Table~\ref{tab:normal-form-num}.

\begin{table}[t]
\caption{Number of different query types of \texttt{[e,p,i,u,n]} system.}
\label{tab:normal-form-num}
\centering
\begin{tabular}{lrrrrr}
\toprule
 & Original & DM & DM+I & DNF & DNF+IU \\ \midrule
Original & 0 & 57 & 132 & 15 & 90 \\
DM & 57 & 0 & 75 & 148 & 256 \\
DM+I & 132 & 75 & 0 & 223 & 181 \\
DNF & 15 & 148 & 223 & 0 & 84 \\
DNF+IU & 90 & 256 & 181 & 84 & 0 \\ \bottomrule
\end{tabular}
\end{table}

\section{Additional Experimental Results}~\label{app:additional_exp}

Here we list detailed experiment results and add some discussions on them.
Due to the large number of the query types, we group and report the queries by the maximum projection length and number of anchor nodes in Table \ref{tab:benchmark}, \ref{FB15k}, and \ref{NELL}. These three tables are the results on FB15k-237, FB15k, NELL, correspondingly .

Our analysis mainly focus on FB15k-237 shown in Table\ref{tab:benchmark}, but FB15k and NELL also shows similar phenomena.

\paragraph{Impact of the Max Projection Length.}
It shows a clear descending trend of performance as the query depth increases in all models. NewLook~\cite{DBLP:conf/kdd/LiuDJZT21} call this phenomenon ``cascading error'' as error propagates in each projection operation. In fact, projection plays a pivotal role in all CQA models. For example, CQD~\cite{DBLP:conf/iclr/ArakelyanDMC21} only trains queries of type \texttt{(p,(e))} and uses logical t-norms to represent union and intersection. Beta~\cite{DBLP:conf/nips/RenL20} doubles training steps for query types only containing projection.~\label{sec:importance of projection} Moreover, the operator of anchor entity is always projection: as queries like \texttt{(u,(e),(e))} or \texttt{(i,(e),(e))} are not allowed, which makes projection be indispensable. For all reasons above, we believe the modeling of projection operation is the fundamental factor to final results.

\paragraph{Impact of the Number of Anchor Nodes.}
A query type with more anchor nodes is not necessarily harder. For example, in depth 2 and 3, queries with 3 anchor nodes have higher scores than those with 2 anchor nodes. This is partially because of the feature of the intersection: we find that \texttt{(I,(p,(e)),(p,(e)),(p,(e)))} has much higher score than \texttt{(I,(p,(e)),(p,(e)))}, and  all of the three models also report similar results in their own experiments. On the contrary, \texttt{(U,(p,(e)),(p,(e)),(p,(e)))} is harder than \texttt{(U,(p,(e)),(p,(e)))} , and other queries with \texttt{U} show similar results. This different behaviour is reasonable, as intersection shrinks answer size while union enlarges it. ~\label{sec:intersection increases performance}The averaged answer size of each group is listed in Table\ref{tab:ans-size-statistics}.

\paragraph{The Choice of Operators.} The NewLook model~\cite{DBLP:conf/kdd/LiuDJZT21} uses difference to replace negation in its query representation, which leads to the fact that DNF+IUd and DNF+IUD are the only two forms that NewLook can fully support. In Table \ref{tab:benchmark}, since all possible query types with one anchor node are \texttt{1p,2p,3p}, the leading performance of NewLook on those queries illustrates that NewLook model projection operation better. However, in more complex query types, LogicE outperforms NewLook, indicating that difference operator \texttt{D/d} might not be a better alternative to negation operator \texttt{n}. Additionally, the multi operator \texttt{D} performs slightly better than the binary \texttt{d}.

In additional, we also report the detailed performances of EPFO and Negation queries on each knowledge graph as categorized by Beta~\cite{DBLP:conf/nips/RenL20}, see Table~16-21. We can see from the results that our benchmark is generally harder and fairer than existing datasets~\cite{DBLP:conf/nips/RenL20}.

\begin{table}[t]
\centering
\caption{Benchmark results(\%) on FB15k. The mark $\dagger$ indicates the query groups that previous datasets have not fully covered. The boldface indicates the best scores. The best scores of the same model are underlined.}
\label{tab:benchmark-fb15k}
\scriptsize
\begin{tabular}{lllrrrrrrrrrr}
\toprule
\multirow{2}{*}{\shortstack[l]{CQA\\ Model}} & \multirow{2}{*}{\shortstack[l]{Normal\\ Form}}        & \multirow{2}{*}{Metric} & \multicolumn{9}{c}{Query type groups (\# anchor nodes, max length of Projection chains)}                                                                       & \multirow{2}{*}{AVG.} \\\cmidrule{4-12}
                             &                                     &                         & (1,1)          & (1,2)          & (1,3)          & (2,1)          & (2,2)$^\dagger$          & (2,3)$^\dagger$          & (3,1$^\dagger$)          & (3,2)$^\dagger$          & (3,3)$^\dagger$          &                                    \\\midrule
\multirow{15}{*}{BetaE} & \multirow{5}{*}{DM} & MRR & 51.86 & 25.48 & 23.55 & 30.24 & 19.14 & 17.46 & 26.69 & 17.73 & 15.54 & 17.18 \\
 &  & HIT@1 & 39.09 & 17.06 & 15.63 & 19.75 & 11.44 & 10.76 & 16.66 & 10.67 & 9.43 & 10.46 \\
 &  & HIT@3 & 59.20 & 27.40 & 25.60 & 35.09 & 21.01 & 18.79 & 31.13 & 19.45 & 16.74 & 18.76 \\
 &  & HIT@10 & 76.20 & 42.36 & 38.82 & 50.50 & 34.32 & 30.53 & 46.44 & 31.65 & 27.42 & 30.30 \\
 &  & RA-Oracle & 47.46 & 33.05 & 32.94 & 33.02 & 24.65 & 24.36 & 29.03 & 21.55 & 20.63 & 21.83 \\ \cmidrule{2-13}
 & \multirow{5}{*}{\shortstack[l]{DM\\+I}} & MRR & 51.86 & 25.48 & 23.55 & 30.24 & 19.14 & 17.46 & 26.78 & 17.78 & 15.60 & 17.22 \\
 &  & HIT@1 & 39.09 & 17.06 & 15.63 & 19.75 & 11.44 & 10.76 & 16.77 & 10.71 & 9.48 & 10.51 \\
 &  & HIT@3 & 59.20 & 27.40 & 25.60 & 35.09 & 21.01 & 18.79 & 31.20 & 19.51 & 16.81 & 18.81 \\
 &  & HIT@10 & 76.20 & 42.36 & 38.82 & 50.50 & 34.32 & 30.53 & 46.51 & 31.70 & 27.48 & 30.35 \\
 &  & RA-Oracle & 47.46 & 33.05 & 32.94 & 33.02 & 24.65 & 24.36 & 29.14 & 21.59 & 20.69 & 21.89 \\  \cmidrule{2-13}
 & \multirow{5}{*}{\shortstack[l]{DNF\\+IU}} & MRR & 51.86 & 25.48 & 23.55 & 38.81 & 21.24 & 19.47 & 33.96 & 20.92 & 18.45 & {\ul 20.31} \\
 &  & HIT@1 & 39.09 & 17.06 & 15.63 & 25.37 & 12.50 & 11.78 & 20.88 & 12.25 & 10.87 & {\ul 12.05} \\
 &  & HIT@3 & 59.20 & 27.40 & 25.60 & 45.48 & 23.30 & 20.89 & 39.69 & 22.89 & 19.76 & {\ul 22.10} \\
 &  & HIT@10 & 76.20 & 42.36 & 38.82 & 65.55 & 38.67 & 34.65 & 60.68 & 38.33 & 33.44 & {\ul 36.74} \\
 &  & RA-Oracle & 47.46 & 33.05 & 32.94 & 44.67 & 28.46 & 28.37 & 39.64 & 27.08 & 26.26 & {\ul 27.51} \\ \midrule
\multirow{15}{*}{LogicE} & \multirow{5}{*}{DM} & MRR & \textbf{61.56} & 29.76 & 25.41 & 38.00 & 22.89 & 20.22 & 32.90 & 21.31 & 18.47 & 20.53 \\
 &  & HIT@1 & \textbf{49.63} & 20.51 & 17.24 & 25.16 & 14.05 & 12.80 & 20.48 & 12.92 & 11.39 & 12.68 \\
 &  & HIT@3 & \textbf{69.51} & 32.56 & 27.71 & 44.89 & 25.56 & 21.99 & 39.10 & 23.74 & 20.12 & 22.71 \\
 &  & HIT@10 & \textbf{82.40} & 47.94 & 41.17 & 62.15 & 40.27 & 34.76 & 56.90 & 37.89 & 32.39 & 35.93 \\
 &  & RA-Oracle & \textbf{58.57} & 37.90 & 35.18 & 43.69 & 29.73 & 28.38 & 37.87 & 26.84 & 25.31 & 26.92 \\ \cmidrule{2-13}
 & \multirow{5}{*}{\shortstack[l]{DM\\+I}} & MRR & \textbf{61.56} & 29.76 & 25.41 & 38.00 & 22.89 & 20.22 & 32.98 & 21.33 & 18.49 & 20.55 \\
 &  & HIT@1 & \textbf{49.63} & 20.51 & 17.24 & 25.16 & 14.05 & 12.80 & 20.56 & 12.95 & 11.40 & 12.70 \\
 &  & HIT@3 & \textbf{69.51} & 32.56 & 27.71 & 44.89 & 25.56 & 21.99 & 39.19 & 23.77 & 20.14 & 22.73 \\
 &  & HIT@10 & \textbf{82.40} & 47.94 & 41.17 & 62.15 & 40.27 & 34.76 & 56.98 & 37.91 & 32.41 & 35.96 \\
 &  & RA-Oracle & \textbf{58.57} & 37.90 & 35.18 & 43.69 & 29.73 & 28.38 & 37.96 & 26.88 & 25.33 & 26.95 \\ \cmidrule{2-13}
 & \multirow{5}{*}{\shortstack[l]{DNF\\+IU}} & MRR & \textbf{61.56} & 29.76 & 25.41 & \textbf{40.83} & \textbf{23.91} & \textbf{21.22} & \textbf{34.92} & \textbf{22.66} & \textbf{19.85} & {\ul \textbf{21.89}} \\
 &  & HIT@1 & \textbf{49.63} & 20.51 & 17.24 & 26.88 & \textbf{14.40} & \textbf{13.15} & \textbf{21.16} & \textbf{13.34} & \textbf{11.86} & {\ul \textbf{13.14}} \\
 &  & HIT@3 & \textbf{69.51} & 32.56 & 27.71 & \textbf{48.51} & \textbf{26.74} & \textbf{23.10} & \textbf{41.63} & \textbf{25.20} & \textbf{21.54} & {\ul \textbf{24.17}} \\
 &  & HIT@10 & \textbf{82.40} & 47.94 & 41.17 & \textbf{67.44} & \textbf{42.82} & \textbf{37.19} & \textbf{62.16} & \textbf{41.37} & \textbf{35.75} & {\ul \textbf{39.33}} \\
 &  & RA-Oracle & \textbf{58.57} & 37.90 & 35.18 & \textbf{47.36} & \textbf{31.56} & \textbf{30.37} & \textbf{40.79} & \textbf{29.16} & \textbf{27.91} & {\ul \textbf{29.38}} \\ \midrule
\multirow{10}{*}{NewLook} & \multirow{5}{*}{\shortstack[l]{DNF\\+IUd}} & MRR & 58.82 & \textbf{30.93} & \textbf{26.12} & \textbf{40.49} & \textbf{22.07} & \textbf{20.69} & \textbf{30.15} & \textbf{19.22} & \textbf{18.48} & \textbf{19.80} \\
 &  & HIT@1 & 46.87 & \textbf{22.02} & \textbf{17.88} & \textbf{28.02} & \textbf{13.55} & \textbf{13.09} & \textbf{17.64} & \textbf{11.16} & \textbf{11.22} & \textbf{11.96} \\
 &  & HIT@3 & 66.32 & \textbf{33.76} & \textbf{28.59} & \textbf{46.29} & \textbf{24.24} & \textbf{22.37} & \textbf{35.06} & \textbf{20.97} & \textbf{19.92} & \textbf{21.58} \\
 &  & HIT@10 & 80.51 & \textbf{48.54} & \textbf{42.01} & \textbf{65.33} & \textbf{38.99} & \textbf{35.60} & \textbf{55.62} & \textbf{35.23} & \textbf{32.70} & \textbf{35.28} \\
 &  & RA-Oracle & 55.31 & \textbf{39.14} & \textbf{36.06} & \textbf{45.99} & \textbf{29.05} & \textbf{29.39} & \textbf{35.25} & \textbf{24.82} & \textbf{25.88} & \textbf{26.57} \\ \cmidrule{2-13}
 & \multirow{5}{*}{\shortstack[l]{DNF\\+IUD}} & MRR & 58.82 & \textbf{30.93} & \textbf{26.12} & 40.49 & 22.07 & 20.69 & 30.42 & 19.35 & 18.51 & {\ul 19.87} \\
 &  & HIT@1 & 46.87 & \textbf{22.02} & \textbf{17.88} & \textbf{28.02} & 13.55 & 13.09 & 17.73 & 11.21 & 11.23 & {\ul 11.99} \\
 &  & HIT@3 & 66.32 & \textbf{33.76} & \textbf{28.59} & 46.29 & 24.24 & 22.37 & 35.48 & 21.14 & 19.94 & {\ul 21.66} \\
 &  & HIT@10 & 80.51 & \textbf{48.54} & \textbf{42.01} & 65.33 & 38.99 & 35.60 & 56.27 & 35.56 & 32.76 & {\ul 35.44} \\
 &  & RA-Oracle & 55.31 & \textbf{39.14} & \textbf{36.06} & 45.99 & 29.05 & 29.39 & 35.54 & 24.99 & 25.92 & {\ul 26.66}\\ \bottomrule
\end{tabular}
\end{table}

\begin{table}[t]
\centering\caption{Benchmark results(\%) on NELL. The mark $\dagger$ indicates the query groups that previous datasets have not fully covered. The boldface indicates the best scores. The best scores of the same model are underlined.}
\label{tab:benchmark-nell}
\scriptsize
\begin{tabular}{lllrrrrrrrrrr}
\toprule
\multirow{2}{*}{\shortstack[l]{CQA\\  Model}} & \multirow{2}{*}{\shortstack[l]{Normal\\ Form}}        & \multirow{2}{*}{Metric} & \multicolumn{9}{c}{Query type groups (\# anchor nodes, max length of Projection chains)}                                                                       & \multirow{2}{*}{AVG.} \\\cmidrule{4-12}
                             &                                     &                         & (1,1)          & (1,2)          & (1,3)          & (2,1)          & (2,2)$^\dagger$          & (2,3)$^\dagger$          & (3,1$^\dagger$)          & (3,2)$^\dagger$          & (3,3)$^\dagger$          &                                    \\\midrule
\multirow{15}{*}{BetaE} & \multirow{5}{*}{DM} & MRR & 28.75 & 10.01 & 9.96 & 13.06 & 8.95 & 8.85 & 8.94 & 8.71 & 8.85 & 8.93 \\
 &  & HIT@1 & 20.41 & 5.87 & 6.15 & 8.25 & 5.35 & 5.47 & 5.23 & 5.39 & 5.60 & 5.58 \\
 &  & HIT@3 & 31.79 & 10.53 & 10.28 & 14.29 & 9.38 & 9.22 & 9.47 & 9.11 & 9.31 & 9.38 \\
 &  & HIT@10 & 45.28 & 17.71 & 17.16 & 22.41 & 15.90 & 15.25 & 16.07 & 14.98 & 15.00 & 15.27 \\
 &  & RA-Oracle & 24.59 & 14.63 & 15.42 & 14.27 & 12.71 & 12.87 & 10.98 & 11.63 & 12.16 & 12.08 \\ \cmidrule{2-13}
 & \multirow{5}{*}{\shortstack[l]{DM\\+I}} & MRR & 28.75 & 10.01 & 9.96 & 13.06 & 8.95 & 8.85 & 8.97 & 8.73 & 8.86 & 8.94 \\
 &  & HIT@1 & 20.41 & 5.87 & 6.15 & 8.25 & 5.35 & 5.47 & 5.25 & 5.40 & 5.60 & 5.59 \\
 &  & HIT@3 & 31.79 & 10.53 & 10.28 & 14.29 & 9.38 & 9.22 & 9.53 & 9.12 & 9.31 & 9.39 \\
 &  & HIT@10 & 45.28 & 17.71 & 17.16 & 22.41 & 15.90 & 15.25 & 16.10 & 15.01 & 15.03 & 15.29 \\
 &  & RA-Oracle & 24.59 & 14.63 & 15.42 & 14.27 & 12.71 & 12.87 & 11.00 & 11.65 & 12.16 & 12.09 \\ \cmidrule{2-13}
 & \multirow{5}{*}{\shortstack[l]{DNF\\+IU}} & MRR & 28.75 & 10.01 & 9.96 & 14.84 & 10.06 & 10.02 & 10.55 & 10.27 & 10.64 & {\ul 10.58} \\
 &  & HIT@1 & 20.41 & 5.87 & 6.15 & 9.22 & 5.96 & 6.13 & 5.99 & 6.25 & 6.63 & {\ul 6.52} \\
 &  & HIT@3 & 31.79 & 10.53 & 10.28 & 16.21 & 10.51 & 10.46 & 11.27 & 10.76 & 11.19 & {\ul 11.12} \\
 &  & HIT@10 & 45.28 & 17.71 & 17.16 & 25.84 & 18.02 & 17.45 & 19.34 & 17.90 & 18.27 & {\ul 18.32} \\
 &  & RA-Oracle & 24.59 & 14.63 & 15.42 & 17.47 & 14.69 & 14.89 & 13.78 & 14.39 & 15.30 & {\ul 14.98} \\ \midrule
\multirow{15}{*}{LogicE} & \multirow{5}{*}{DM} & MRR & \textbf{35.23} & \textbf{13.53} & \textbf{13.75} & 16.16 & 11.40 & 11.35 & 11.06 & 10.75 & 11.05 & 11.13 \\
 &  & HIT@1 & \textbf{25.74} & \textbf{8.62} & \textbf{8.92} & 10.32 & 7.20 & 7.52 & 6.44 & 6.82 & 7.36 & 7.26 \\
 &  & HIT@3 & \textbf{40.27} & \textbf{14.28} & \textbf{14.73} & 17.85 & 12.16 & 12.02 & 11.97 & 11.45 & 11.79 & 11.89 \\
 &  & HIT@10 & \textbf{53.28} & \textbf{23.05} & \textbf{22.80} & 27.52 & 19.24 & 18.52 & 19.97 & 18.09 & 17.94 & 18.38 \\
 &  & RA-Oracle & \textbf{30.69} & \textbf{19.62} & \textbf{20.32} & 18.94 & 16.33 & 16.16 & 14.41 & 14.77 & 15.23 & 15.25 \\ \cmidrule{2-13}
 & \multirow{5}{*}{\shortstack[l]{DM\\+I}} & MRR & \textbf{35.23} & \textbf{13.53} & \textbf{13.75} & 16.16 & 11.40 & 11.35 & 11.07 & 10.75 & 11.06 & 11.14 \\
 &  & HIT@1 & \textbf{25.74} & \textbf{8.62} & \textbf{8.92} & 10.32 & 7.20 & 7.52 & 6.49 & 6.84 & 7.37 & 7.27 \\
 &  & HIT@3 & \textbf{40.27} & \textbf{14.28} & \textbf{14.73} & 17.85 & 12.16 & 12.02 & 11.95 & 11.44 & 11.80 & 11.89 \\
 &  & HIT@10 & \textbf{53.28} & \textbf{23.05} & \textbf{22.80} & 27.52 & 19.24 & 18.52 & 19.96 & 18.10 & 17.95 & 18.39 \\
 &  & RA-Oracle & \textbf{30.69} & \textbf{19.62} & \textbf{20.32} & 18.94 & 16.33 & 16.16 & 14.40 & 14.77 & 15.23 & 15.26 \\ \cmidrule{2-13}
 & \multirow{5}{*}{\shortstack[l]{DNF\\+IU}} & MRR & \textbf{35.23} & \textbf{13.53} & \textbf{13.75} & \textbf{18.19} & \textbf{12.94} & \textbf{13.00} & \textbf{12.40} & \textbf{12.48} & \textbf{13.22} & {\ul \textbf{13.07}} \\
 &  & HIT@1 & \textbf{25.74} & \textbf{8.62} & \textbf{8.92} & \textbf{11.45} & \textbf{8.05} & \textbf{8.42} & \textbf{6.98} & \textbf{7.71} & \textbf{8.57} & {\ul \textbf{8.31}} \\
 &  & HIT@3 & \textbf{40.27} & \textbf{14.28} & \textbf{14.73} & \textbf{20.13} & \textbf{13.90} & \textbf{13.86} & \textbf{13.41} & \textbf{13.33} & \textbf{14.16} & {\ul \textbf{14.01}} \\
 &  & HIT@10 & \textbf{53.28} & \textbf{23.05} & \textbf{22.80} & \textbf{31.35} & \textbf{22.15} & \textbf{21.61} & \textbf{23.01} & \textbf{21.49} & \textbf{21.94} & {\ul \textbf{22.04}} \\
 &  & RA-Oracle & \textbf{30.69} & \textbf{19.62} & \textbf{20.32} & \textbf{22.08} & \textbf{18.84} & \textbf{18.87} & \textbf{16.62} & \textbf{17.55} & \textbf{18.75} & {\ul \textbf{18.39}} \\ \midrule
\multirow{10}{*}{NewLook} & \multirow{5}{*}{\shortstack[l]{DNF\\+IUd}} & MRR & 33.59 & 12.06 & 11.42 & 16.55 & 10.30 & 10.52 & 9.50 & 8.91 & 10.08 & 9.88 \\
 &  & HIT@1 & 24.78 & 7.12 & 6.79 & 10.59 & 6.22 & 6.47 & 5.33 & 5.28 & 6.25 & {\ul 6.04} \\
 &  & HIT@3 & 37.19 & 12.73 & 12.00 & 18.00 & 10.82 & 10.99 & 10.00 & 9.31 & 10.53 & 10.35 \\
 &  & HIT@10 & 51.89 & 21.72 & 20.31 & 28.47 & 17.98 & 18.12 & 17.63 & 15.70 & 17.24 & 17.10 \\
 &  & RA-Oracle & 29.50 & 18.03 & 17.76 & 19.27 & 15.07 & 15.85 & 12.21 & 12.50 & 14.76 & 14.15 \\ \cmidrule{2-13}
 & \multirow{5}{*}{\shortstack[l]{DNF\\+IUD}} & MRR & 33.59 & 12.06 & 11.42 & 16.55 & 10.30 & 10.52 & 9.53 & 8.94 & 10.09 & {\ul 9.90} \\
 &  & HIT@1 & 24.78 & 7.12 & 6.79 & 10.59 & 6.22 & 6.47 & 5.34 & 5.29 & 6.26 & {\ul 6.04} \\
 &  & HIT@3 & 37.19 & 12.73 & 12.00 & 18.00 & 10.82 & 10.99 & 10.05 & 9.34 & 10.54 & {\ul 10.36} \\
 &  & HIT@10 & 51.89 & 21.72 & 20.31 & 28.47 & 17.98 & 18.12 & 17.68 & 15.75 & 17.25 & {\ul 17.13} \\
 &  & RA-Oracle & 29.50 & 18.03 & 17.76 & 19.27 & 15.07 & 15.85 & 12.24 & 12.53 & 14.76 & {\ul 14.16} \\ \bottomrule
\end{tabular}
\end{table}

\begin{table}[t]
\centering
\caption{Average number of answers of queries.}
\label{tab:ans-size-statistics}
\begin{tabular}{llllllllll}
\toprule
\multicolumn{1}{c}{\multirow{2}{*}{Knowledge   Graph}} & \multicolumn{9}{c}{Query type groups (\# anchor nodes, max length of Projection chains)} \\ \cmidrule{2-10}
\multicolumn{1}{c}{} & \multicolumn{1}{c}{(1,1)} & \multicolumn{1}{c}{(1,2)} & \multicolumn{1}{c}{(1,3)} & \multicolumn{1}{c}{(2,1)} & \multicolumn{1}{c}{(2,2)} & \multicolumn{1}{c}{(2,3)} & \multicolumn{1}{c}{(3,1)} & \multicolumn{1}{c}{(3,2)} & \multicolumn{1}{c}{(3,3)} \\ \midrule
FB15k-237 & 3.22 & 17.56 & 23.44 & 12.37 & 18.49 & 22.61 & 13.16 & 17.60 & 21.22 \\
FB15k & 4.09 & 20.07 & 24.26 & 15.34 & 20.42 & 24.02 & 15.38 & 19.37 & 23.02 \\
NELL & 3.13 & 20.54 & 22.23 & 16.66 & 22.08 & 22.87 & 19.22 & 21.18 & 22.84 \\ \bottomrule
\end{tabular}
\end{table}

\begin{table}[t]
\centering
\scriptsize
\caption{Benchmark results(\%) on EPFO queries of FB15k-237.}
\begin{tabular}{lllrrrrrrrrrr}
\toprule
\multirow{2}{*}{\shortstack[l]{CQA\\  Model}} & \multirow{2}{*}{\shortstack[l]{Normal\\ Form}} & \multirow{2}{*}{Metric} & \multicolumn{9}{c}{(\# anchor nodes, max   length of projection chains)} & \multicolumn{1}{c}{\multirow{2}{*}{AVG.}} \\ \cmidrule{4-12}
 &  & \multicolumn{1}{c}{} & \multicolumn{1}{c}{(1,1)} & \multicolumn{1}{c}{(1,2)} & \multicolumn{1}{c}{(1,3)} & \multicolumn{1}{c}{(2,1)} & \multicolumn{1}{c}{(2,2)} & \multicolumn{1}{c}{(2,3)} & \multicolumn{1}{c}{(3,1)} & \multicolumn{1}{c}{(3,2)} & \multicolumn{1}{c}{(3,3)} & \multicolumn{1}{c}{} \\ \midrule
\multirow{15}{*}{BetaE} & \multirow{5}{*}{DM} & MRR & 18.79 & 9.72 & 9.64 & 15.51 & 10.00 & 8.78 & 15.05 & 11.09 & 9.33 & 9.98 \\
 &  & HIT@1 & 10.63 & 4.63 & 4.68 & 9.53 & 5.39 & 4.44 & 9.71 & 6.60 & 5.16 & 5.63 \\
 &  & HIT@3 & 20.37 & 9.61 & 9.44 & 16.97 & 10.35 & 8.88 & 16.52 & 11.67 & 9.58 & 10.35 \\
 &  & HIT@10 & 36.19 & 19.80 & 19.38 & 27.44 & 18.96 & 16.96 & 25.45 & 19.72 & 17.23 & 18.30 \\
 &  & RA-Oracle & 14.38 & 14.40 & 16.99 & 16.00 & 13.47 & 13.66 & 15.11 & 12.99 & 12.57 & 12.93 \\ \cmidrule{2-13}
 & \multirow{5}{*}{\shortstack[l]{DM\\+I}} & MRR & 18.79 & 9.72 & 9.64 & 15.51 & 10.00 & 8.78 & 15.19 & 11.12 & 9.37 & 10.02 \\
 &  & HIT@1 & 10.63 & 4.63 & 4.68 & 9.53 & 5.39 & 4.44 & 9.92 & 6.62 & 5.20 & 5.66 \\
 &  & HIT@3 & 20.37 & 9.61 & 9.44 & 16.97 & 10.35 & 8.88 & 16.58 & 11.72 & 9.64 & 10.40 \\
 &  & HIT@10 & 36.19 & 19.80 & 19.38 & 27.44 & 18.96 & 16.96 & 25.52 & 19.77 & 17.29 & 18.35 \\
 &  & RA-Oracle & 14.38 & 14.40 & 16.99 & 16.00 & 13.47 & 13.66 & 15.32 & 13.03 & 12.63 & 12.98 \\ \cmidrule{2-13}
 & \multirow{5}{*}{\shortstack[l]{DNF\\+IU}} & MRR & 18.79 & 9.72 & 9.64 & 17.97 & 11.35 & 9.77 & 19.11 & 13.41 & 11.05 & 11.81 \\
 &  & HIT@1 & 10.63 & 4.63 & 4.68 & 10.59 & 5.96 & 4.85 & 11.85 & 7.72 & 5.99 & 6.50 \\
 &  & HIT@3 & 20.37 & 9.61 & 9.44 & 19.43 & 11.59 & 9.82 & 20.86 & 14.01 & 11.29 & 12.17 \\
 &  & HIT@10 & 36.19 & 19.80 & 19.38 & 33.10 & 21.84 & 19.08 & 33.62 & 24.46 & 20.70 & 22.01 \\
 &  & RA-Oracle & 14.38 & 14.40 & 16.99 & 20.17 & 15.98 & 15.81 & 21.34 & 16.90 & 15.79 & 16.23 \\  \midrule
\multirow{15}{*}{LogicE} & \multirow{5}{*}{DM} & MRR & 20.71 & 10.70 & 10.18 & 19.49 & 12.16 & 10.35 & 19.07 & 13.50 & 11.16 & 12.00 \\
 &  & HIT@1 & 11.66 & 5.20 & 5.25 & 12.07 & 6.68 & 5.56 & 12.38 & 8.03 & 6.34 & 6.91 \\
 &  & HIT@3 & 23.02 & 10.66 & 9.96 & 21.36 & 12.62 & 10.55 & 20.65 & 14.32 & 11.55 & 12.53 \\
 &  & HIT@10 & 39.81 & 21.25 & 19.48 & 34.59 & 22.86 & 19.51 & 32.51 & 24.13 & 20.43 & 21.86 \\
 &  & RA-Oracle & 15.64 & 15.27 & 17.28 & 21.13 & 16.49 & 15.82 & 20.77 & 16.62 & 15.37 & 15.94 \\ \cmidrule{2-13}
 & \multirow{5}{*}{\shortstack[l]{DM\\+I}} & MRR & 20.71 & 10.70 & 10.18 & 19.49 & 12.16 & 10.35 & 19.13 & 13.52 & 11.17 & 12.02 \\
 &  & HIT@1 & 11.66 & 5.20 & 5.25 & 12.07 & 6.68 & 5.56 & 12.42 & 8.04 & 6.34 & 6.91 \\
 &  & HIT@3 & 23.02 & 10.66 & 9.96 & 21.36 & 12.62 & 10.55 & 20.81 & 14.34 & 11.58 & 12.56 \\
 &  & HIT@10 & 39.81 & 21.25 & 19.48 & 34.59 & 22.86 & 19.51 & 32.62 & 24.16 & 20.45 & 21.89 \\
 &  & RA-Oracle & 15.64 & 15.27 & 17.28 & 21.13 & 16.49 & 15.82 & 20.79 & 16.66 & 15.39 & 15.95 \\ \cmidrule{2-13}
 & \multirow{5}{*}{\shortstack[l]{DNF\\+IU}} & MRR & 20.71 & 10.70 & 10.18 & 19.79 & 12.59 & 10.68 & 20.21 & 14.50 & \textbf{11.95} & 12.78 \\
 &  & HIT@1 & 11.66 & 5.20 & \textbf{5.25} & 11.90 & 6.79 & 5.62 & 12.67 & 8.44 & 6.68 & 7.22 \\
 &  & HIT@3 & 23.02 & 10.66 & 9.96 & 21.56 & 13.03 & \textbf{10.86} & 21.97 & 15.31 & \textbf{12.34} & 13.31 \\
 &  & HIT@10 & 39.81 & 21.25 & 19.48 & 36.03 & \textbf{23.85} & \textbf{20.27} & 35.48 & \textbf{26.30} & \textbf{21.99} & \textbf{23.48} \\
 &  & RA-Oracle & 15.64 & 15.27 & 17.28 & 21.81 & 17.31 & 16.55 & 22.04 & 18.08 & 16.64 & 17.16 \\ \midrule
\multirow{5}{*}{NewLook} & \multirow{5}{*}{\shortstack[l]{DNF\\+IUD}} & MRR & \textbf{22.31} & \textbf{11.19} & \textbf{10.39} & \textbf{21.66} & \textbf{12.76} & \textbf{10.77} & \textbf{21.76} & \textbf{14.74} & \textbf{11.95} & \textbf{12.92} \\
 &  & HIT@1 & \textbf{13.55} & \textbf{5.62} & 5.18 & \textbf{13.74} & \textbf{7.09} & \textbf{5.85} & \textbf{14.24} & \textbf{8.88} & \textbf{6.84} & \textbf{7.52} \\
 &  & HIT@3 & \textbf{24.62} & \textbf{11.40} & \textbf{10.38} & \textbf{23.85} & \textbf{13.22} & 10.83 & \textbf{23.66} & \textbf{15.49} & 12.27 & \textbf{13.40} \\
 &  & HIT@10 & \textbf{40.53} & \textbf{22.18} & \textbf{20.47} & \textbf{37.48} & 23.66 & 20.04 & \textbf{36.82} & 26.06 & 21.66 & 23.28 \\
 &  & RA-Oracle & \textbf{17.66} & \textbf{16.32} & \textbf{17.79} & \textbf{24.11} & \textbf{17.73} & \textbf{16.85} & \textbf{24.15} & \textbf{18.63} & \textbf{16.92} & \textbf{17.59} \\ \bottomrule
\end{tabular}
\end{table}

\begin{table}[t]
\centering
\scriptsize
\caption{Benchmark results(\%) on queries with negation of FB15k-237.}
\begin{tabular}{lllrrrrrrr}
\toprule
\multirow{2}{*}{\shortstack[l]{CQA\\  Model}} & \multirow{2}{*}{\shortstack[l]{Normal\\ Form}} & \multirow{2}{*}{Metric} & \multicolumn{6}{c}{(\# anchor nodes, max   length of projection chains)} & \multicolumn{1}{c}{\multirow{2}{*}{AVG.}} \\ \cmidrule{4-9}
 &  & \multicolumn{1}{c}{} & \multicolumn{1}{c}{(2,1)} & \multicolumn{1}{c}{(2,2)} & \multicolumn{1}{c}{(2,3)} & \multicolumn{1}{c}{(3,1)} & \multicolumn{1}{c}{(3,2)} & \multicolumn{1}{c}{(3,3)} & \multicolumn{1}{c}{} \\ \midrule
\multirow{15}{*}{BetaE} & \multirow{5}{*}{DM} & MRR & 7.24 & 6.19 & 5.84 & 9.49 & 7.22 & 6.47 & 6.92 \\
 &  & HIT@1 & 2.15 & 2.25 & 2.03 & 4.14 & 3.24 & 2.85 & 3.04 \\
 &  & HIT@3 & 6.47 & 5.41 & 5.14 & 9.72 & 7.02 & 6.17 & 6.66 \\
 &  & HIT@10 & 17.92 & 13.61 & 12.96 & 20.26 & 14.61 & 13.07 & 14.12 \\
 &  & RA-Oracle & 10.26 & 9.99 & 10.95 & 11.21 & 9.70 & 10.04 & 9.99 \\ \cmidrule{2-10}
 & \multirow{5}{*}{\shortstack[l]{DM\\+I}} & MRR & 7.24 & 6.19 & 5.84 & 9.48 & 7.22 & 6.47 & 6.92 \\
 &  & HIT@1 & 2.15 & 2.25 & 2.03 & 4.12 & 3.24 & 2.85 & 3.04 \\
 &  & HIT@3 & 6.47 & 5.41 & 5.14 & 9.73 & 7.03 & 6.17 & 6.67 \\
 &  & HIT@10 & 17.92 & 13.61 & 12.96 & 20.26 & 14.62 & 13.08 & 14.12 \\
 &  & RA-Oracle & 10.26 & 9.99 & 10.95 & 11.20 & 9.71 & 10.05 & 9.99 \\ \cmidrule{2-10}
 & \multirow{5}{*}{\shortstack[l]{DNF\\+IU}} & MRR & 7.24 & 6.19 & 5.84 & 10.16 & 7.78 & 7.04 & 7.46 \\
 &  & HIT@1 & 2.15 & 2.25 & 2.03 & 4.32 & 3.42 & 3.04 & 3.21 \\
 &  & HIT@3 & 6.47 & 5.41 & 5.14 & 10.36 & 7.50 & 6.64 & 7.12 \\
 &  & HIT@10 & 17.92 & 13.61 & 12.96 & 21.92 & 15.88 & 14.32 & 15.33 \\
 &  & RA-Oracle & 10.26 & 9.99 & 10.95 & 12.41 & 10.78 & 11.25 & 11.08 \\ \midrule
\multirow{15}{*}{LogicE} & \multirow{5}{*}{\shortstack[l]{DM\\+I}} & MRR & \textbf{8.01} & \textbf{6.80} & \textbf{6.28} & 11.03 & 8.29 & 7.42 & 7.93 \\
 &  & HIT@1 & \textbf{2.28} & \textbf{2.48} & \textbf{2.39} & 4.88 & 3.77 & 3.38 & 3.57 \\
 &  & HIT@3 & \textbf{7.44} & \textbf{6.23} & \textbf{5.67} & 11.54 & 8.17 & 7.19 & 7.76 \\
 &  & HIT@10 & \textbf{19.80} & \textbf{14.87} & \textbf{13.48} & 23.24 & 16.81 & 14.89 & 16.11 \\
 &  & RA-Oracle & \textbf{10.21} & \textbf{10.18} & \textbf{11.19} & 13.04 & 11.02 & 11.30 & 11.24 \\ \cmidrule{2-10}
 & \multirow{5}{*}{DM+I} & MRR & \textbf{8.01} & \textbf{6.80} & \textbf{6.28} & \textbf{11.07} & 8.31 & 7.44 & 7.94 \\
 &  & HIT@1 & \textbf{2.28} & \textbf{2.48} & \textbf{2.39} & \textbf{4.90} & \textbf{3.79} & 3.39 & \textbf{3.58} \\
 &  & HIT@3 & \textbf{7.44} & \textbf{6.23} & \textbf{5.67} & \textbf{11.59} & 8.19 & 7.22 & 7.79 \\
 &  & HIT@10 & \textbf{19.80} & \textbf{14.87} & \textbf{13.48} & 23.31 & 16.85 & 14.92 & 16.14 \\
 &  & RA-Oracle & \textbf{10.21} & \textbf{10.18} & \textbf{11.19} & \textbf{13.06} & 11.04 & 11.32 & 11.26 \\ \cmidrule{2-10}
 & \multirow{5}{*}{\shortstack[l]{DNF\\+IU}} & MRR & \textbf{8.01} & \textbf{6.80} & \textbf{6.28} & 10.99 & \textbf{8.42} & \textbf{7.58} & \textbf{8.05} \\
 &  & HIT@1 & \textbf{2.28} & \textbf{2.48} & \textbf{2.39} & 4.70 & 3.77 & \textbf{3.41} & 3.57 \\
 &  & HIT@3 & \textbf{7.44} & \textbf{6.23} & \textbf{5.67} & 11.37 & \textbf{8.22} & \textbf{7.30} & \textbf{7.82} \\
 &  & HIT@10 & \textbf{19.80} & \textbf{14.87} & \textbf{13.48} & \textbf{23.48} & \textbf{17.18} & \textbf{15.20} & \textbf{16.42} \\
 &  & RA-Oracle & \textbf{10.21} & \textbf{10.18} & \textbf{11.19} & 12.96 & \textbf{11.23} & \textbf{11.63} & \textbf{11.49} \\ \midrule
\multirow{10}{*}{NewLook} & \multirow{5}{*}{\shortstack[l]{DNF\\+IUd}} & MRR & 4.73 & 4.50 & 4.38 & 6.43 & 5.30 & 5.00 & 5.17 \\
 &  & HIT@1 & 0.80 & 1.48 & 1.50 & 2.14 & 2.08 & 1.97 & 1.99 \\
 &  & HIT@3 & 4.25 & 3.77 & 3.71 & 6.21 & 4.90 & 4.54 & 4.74 \\
 &  & HIT@10 & 12.34 & 10.01 & 9.35 & 14.91 & 11.06 & 10.32 & 10.85 \\
 &  & RA-Oracle & 4.37 & 5.90 & 7.36 & 6.52 & 6.63 & 7.65 & 7.10 \\ \cmidrule{2-10}
 & \multirow{5}{*}{\shortstack[l]{DNF\\+IUD}} & MRR & 4.73 & 4.50 & 4.38 & 6.50 & 5.35 & 5.02 & 5.20 \\
 &  & HIT@1 & 0.80 & 1.48 & 1.50 & 2.16 & 2.10 & 1.97 & 2.00 \\
 &  & HIT@3 & 4.25 & 3.77 & 3.71 & 6.26 & 4.95 & 4.56 & 4.76 \\
 &  & HIT@10 & 12.34 & 10.01 & 9.35 & 15.08 & 11.18 & 10.35 & 10.92 \\
 &  & RA-Oracle & 4.37 & 5.90 & 7.36 & 6.57 & 6.67 & 7.67 & 7.14 \\ \bottomrule
\end{tabular}
\end{table}

\begin{table}[t]
\centering
\scriptsize
\caption{Benchmark results(\%) on EPFO queries of FB15k.}
\begin{tabular}{lllrrrrrrrrrr}
\toprule
\multirow{2}{*}{\shortstack[l]{CQA\\  Model}} & \multirow{2}{*}{\shortstack[l]{Normal\\ Form}} & \multirow{2}{*}{Metric} & \multicolumn{9}{c}{(\# anchor nodes, max   length of projection chains)} & \multicolumn{1}{c}{\multirow{2}{*}{AVG.}} \\ \cmidrule{4-12}
 &  & \multicolumn{1}{c}{} & \multicolumn{1}{c}{(1,1)} & \multicolumn{1}{c}{(1,2)} & \multicolumn{1}{c}{(1,3)} & \multicolumn{1}{c}{(2,1)} & \multicolumn{1}{c}{(2,2)} & \multicolumn{1}{c}{(2,3)} & \multicolumn{1}{c}{(3,1)} & \multicolumn{1}{c}{(3,2)} & \multicolumn{1}{c}{(3,3)} & \multicolumn{1}{c}{} \\ \midrule
\multirow{15}{*}{BetaE} & \multirow{5}{*}{DM} & MRR & 51.86 & 25.48 & 23.55 & 32.07 & 20.42 & 18.66 & 28.51 & 20.26 & 17.36 & 18.97 \\
 &  & HIT@1 & 39.09 & 17.06 & 15.63 & 23.82 & 13.37 & 12.02 & 21.38 & 13.74 & 11.22 & 12.56 \\
 &  & HIT@3 & 59.20 & 27.40 & 25.60 & 35.86 & 22.30 & 20.17 & 31.63 & 22.24 & 18.88 & 20.72 \\
 &  & HIT@10 & 76.20 & 42.36 & 38.82 & 47.78 & 34.22 & 31.56 & 42.08 & 32.91 & 29.27 & 31.41 \\
 &  & RA-Oracle & 47.46 & 33.05 & 32.94 & 33.80 & 25.61 & 25.21 & 29.44 & 23.42 & 22.08 & 23.36 \\ \cmidrule{2-13}
 & \multirow{5}{*}{\shortstack[l]{DM\\+I}} & MRR & 51.86 & 25.48 & 23.55 & 32.07 & 20.42 & 18.66 & 28.76 & 20.38 & 17.46 & 19.07 \\
 &  & HIT@1 & 39.09 & 17.06 & 15.63 & 23.82 & 13.37 & 12.02 & 21.69 & 13.85 & 11.31 & 12.65 \\
 &  & HIT@3 & 59.20 & 27.40 & 25.60 & 35.86 & 22.30 & 20.17 & 31.84 & 22.39 & 18.99 & 20.83 \\
 &  & HIT@10 & 76.20 & 42.36 & 38.82 & 47.78 & 34.22 & 31.56 & 42.24 & 33.04 & 29.38 & 31.51 \\
 &  & RA-Oracle & 47.46 & 33.05 & 32.94 & 33.80 & 25.61 & 25.21 & 29.74 & 23.55 & 22.19 & 23.46 \\ \cmidrule{2-13}
 & \multirow{5}{*}{\shortstack[l]{DNF\\+IU}} & MRR & 51.86 & 25.48 & 23.55 & 44.93 & 23.92 & 21.26 & 44.04 & 26.31 & 21.48 & 23.71 \\
 &  & HIT@1 & 39.09 & 17.06 & 15.63 & 32.24 & 15.14 & 13.34 & 31.69 & 17.14 & 13.40 & 15.18 \\
 &  & HIT@3 & 59.20 & 27.40 & 25.60 & 51.44 & 26.11 & 22.91 & 49.79 & 28.88 & 23.22 & 25.84 \\
 &  & HIT@10 & 76.20 & 42.36 & 38.82 & 70.36 & 41.48 & 36.92 & 69.08 & 44.66 & 37.49 & 40.66 \\
 &  & RA-Oracle & 47.46 & 33.05 & 32.94 & 51.28 & 31.96 & 30.44 & 51.04 & 33.54 & 29.74 & 31.67 \\ \midrule
\multirow{15}{*}{LogicE} & \multirow{5}{*}{DM} & MRR & \textbf{61.56} & 29.76 & 25.41 & 42.92 & 25.26 & 21.77 & 39.51 & 25.46 & 21.04 & 23.30 \\
 &  & HIT@1 & \textbf{49.63} & 20.51 & 17.24 & 31.77 & 16.89 & 14.30 & 29.03 & 17.30 & 13.79 & 15.61 \\
 &  & HIT@3 & \textbf{69.51} & 32.56 & 27.71 & 48.83 & 27.88 & 23.75 & 44.87 & 28.24 & 23.06 & 25.68 \\
 &  & HIT@10 & \textbf{82.40} & 47.94 & 41.17 & 63.89 & 41.68 & 36.38 & 59.37 & 41.32 & 35.19 & 38.24 \\
 &  & RA-Oracle & \textbf{58.57} & 37.90 & 35.18 & 48.91 & 32.53 & 29.93 & 44.83 & 31.38 & 27.97 & 30.02 \\ \cmidrule{2-13}
 & \multirow{5}{*}{\shortstack[l]{DM\\+I}} & MRR & \textbf{61.56} & 29.76 & 25.41 & 42.92 & 25.26 & 21.77 & 39.63 & 25.47 & 21.05 & 23.31 \\
 &  & HIT@1 & \textbf{49.63} & 20.51 & 17.24 & 31.77 & 16.89 & 14.30 & 29.16 & 17.32 & 13.80 & 15.63 \\
 &  & HIT@3 & \textbf{69.51} & 32.56 & 27.71 & 48.83 & 27.88 & 23.75 & 44.94 & 28.24 & 23.07 & 25.69 \\
 &  & HIT@10 & \textbf{82.40} & 47.94 & 41.17 & 63.89 & 41.68 & 36.38 & 59.50 & 41.35 & 35.21 & 38.27 \\
 &  & RA-Oracle & \textbf{58.57} & 37.90 & 35.18 & 48.91 & 32.53 & 29.93 & 44.98 & 31.42 & 27.99 & 30.04 \\ \cmidrule{2-13}
 & \multirow{5}{*}{\shortstack[l]{DNF\\+IU}} & MRR & \textbf{61.56} & 29.76 & 25.41 & 47.17 & 26.96 & 23.08 & 44.33 & 28.13 & 23.01 & 25.42 \\
 &  & HIT@1 & \textbf{49.63} & 20.51 & 17.24 & 34.35 & 17.47 & 14.76 & 31.62 & 18.42 & 14.55 & 16.48 \\
 &  & HIT@3 & \textbf{69.51} & 32.56 & 27.71 & 54.27 & 29.86 & 25.19 & 50.68 & 31.22 & 25.15 & 28.01 \\
 &  & HIT@10 & \textbf{82.40} & 47.94 & 41.17 & 71.83 & 45.92 & 39.54 & 69.46 & 47.54 & 39.82 & 43.19 \\
 &  & RA-Oracle & \textbf{58.57} & 37.90 & 35.18 & 54.42 & 35.59 & 32.52 & 51.76 & 35.88 & 31.59 & 33.77 \\ \midrule
\multirow{5}{*}{NewLook} & \multirow{5}{*}{\shortstack[l]{DNF\\+IUD}} & MRR & 58.82 & \textbf{30.93} & \textbf{26.12} & \textbf{52.60} & \textbf{28.84} & \textbf{24.06} & \textbf{50.64} & \textbf{30.87} & \textbf{24.51} & \textbf{27.31} \\
 &  & HIT@1 & 46.87 & \textbf{22.02} & \textbf{17.88} & \textbf{40.51} & \textbf{19.61} & \textbf{15.92} & \textbf{38.06} & \textbf{21.24} & \textbf{16.22} & \textbf{18.52} \\
 &  & HIT@3 & 66.32 & \textbf{33.76} & \textbf{28.59} & \textbf{59.44} & \textbf{31.81} & \textbf{26.21} & \textbf{57.66} & \textbf{34.30} & \textbf{26.81} & \textbf{30.08} \\
 &  & HIT@10 & 80.51 & \textbf{48.54} & \textbf{42.01} & \textbf{75.89} & \textbf{47.17} & \textbf{40.03} & \textbf{75.04} & \textbf{49.94} & \textbf{40.86} & \textbf{44.62} \\
 &  & RA-Oracle & 55.31 & \textbf{39.14} & \textbf{36.06} & \textbf{58.44} & \textbf{37.32} & \textbf{33.49} & \textbf{57.14} & \textbf{38.70} & \textbf{33.14} & \textbf{35.65} \\ \bottomrule
\end{tabular}
\end{table}

\begin{table}[t]
\centering
\scriptsize
\caption{Benchmark results(\%) on queries with negation of FB15k.}
\begin{tabular}{lllrrrrrrr}
\toprule
\multirow{2}{*}{\shortstack[l]{CQA\\  Model}} & \multirow{2}{*}{\shortstack[l]{Normal\\ Form}} & \multirow{2}{*}{Metric} & \multicolumn{6}{c}{(\# anchor nodes, max   length of projection chains)} & \multicolumn{1}{c}{\multirow{2}{*}{AVG.}} \\ \cmidrule{4-9}
 &  & \multicolumn{1}{c}{} & \multicolumn{1}{c}{(2,1)} & \multicolumn{1}{c}{(2,2)} & \multicolumn{1}{c}{(2,3)} & \multicolumn{1}{c}{(3,1)} & \multicolumn{1}{c}{(3,2)} & \multicolumn{1}{c}{(3,3)} & \multicolumn{1}{c}{} \\ \midrule
\multirow{15}{*}{BetaE} & \multirow{5}{*}{DM} & MRR & 26.59 & 17.23 & 13.48 & 25.77 & 16.36 & 13.14 & 15.31 \\
 &  & HIT@1 & 11.62 & 8.54 & 6.58 & 14.29 & 9.00 & 7.07 & 8.29 \\
 &  & HIT@3 & 33.56 & 19.08 & 14.17 & 30.88 & 17.94 & 13.93 & 16.73 \\
 &  & HIT@10 & 55.93 & 34.46 & 27.09 & 48.63 & 30.97 & 24.98 & 29.15 \\
 &  & RA-Oracle & 31.46 & 23.21 & 21.50 & 28.83 & 20.53 & 18.73 & 20.26 \\ \cmidrule{2-10}
 & \multirow{5}{*}{\shortstack[l]{DM\\+I}} & MRR & 26.59 & 17.23 & 13.48 & 25.78 & 16.36 & 13.14 & 15.32 \\
 &  & HIT@1 & 11.62 & 8.54 & 6.58 & 14.31 & 9.00 & 7.07 & 8.29 \\
 &  & HIT@3 & 33.56 & 19.08 & 14.17 & 30.88 & 17.94 & 13.93 & 16.72 \\
 &  & HIT@10 & 55.93 & 34.46 & 27.09 & 48.64 & 30.97 & 24.98 & 29.15 \\
 &  & RA-Oracle & 31.46 & 23.21 & 21.50 & 28.83 & 20.53 & 18.72 & 20.26 \\ \cmidrule{2-10}
 & \multirow{5}{*}{\shortstack[l]{DNF\\+IU}} & MRR & 26.59 & 17.23 & 13.48 & 28.92 & 18.00 & 14.46 & 16.79 \\
 &  & HIT@1 & 11.62 & 8.54 & 6.58 & 15.48 & 9.60 & 7.53 & 8.82 \\
 &  & HIT@3 & 33.56 & 19.08 & 14.17 & 34.64 & 19.65 & 15.20 & 18.23 \\
 &  & HIT@10 & 55.93 & 34.46 & 27.09 & 56.48 & 34.90 & 28.11 & 32.69 \\
 &  & RA-Oracle & 31.46 & 23.21 & 21.50 & 33.95 & 23.57 & 21.69 & 23.21 \\ \midrule
\multirow{15}{*}{LogicE} & \multirow{5}{*}{DM} & MRR & \textbf{28.17} & \textbf{19.34} & \textbf{15.04} & 29.59 & 19.06 & 15.10 & 17.66 \\
 &  & HIT@1 & \textbf{11.94} & \textbf{9.80} & \textbf{7.80} & 16.20 & 10.54 & 8.25 & 9.65 \\
 &  & HIT@3 & \textbf{37.01} & \textbf{22.08} & \textbf{16.13} & 36.21 & 21.30 & 16.26 & 19.64 \\
 &  & HIT@10 & \textbf{58.66} & \textbf{38.16} & \textbf{29.35} & 55.66 & 36.02 & 28.70 & 33.55 \\
 &  & RA-Oracle & \textbf{33.25} & \textbf{25.52} & \textbf{23.21} & 34.38 & 24.38 & 21.81 & 23.72 \\ \cmidrule{2-10}
 & \multirow{5}{*}{\shortstack[l]{DM\\+I}} & MRR & \textbf{28.17} & \textbf{19.34} & \textbf{15.04} & 29.65 & 19.09 & 15.12 & 17.69 \\
 &  & HIT@1 & \textbf{11.94} & \textbf{9.80} & \textbf{7.80} & \textbf{16.25} & 10.57 & 8.26 & 9.67 \\
 &  & HIT@3 & \textbf{37.01} & \textbf{22.08} & \textbf{16.13} & 36.31 & 21.34 & 16.29 & 19.68 \\
 &  & HIT@10 & \textbf{58.66} & \textbf{38.16} & \textbf{29.35} & 55.71 & 36.04 & 28.73 & 33.57 \\
 &  & RA-Oracle & \textbf{33.25} & \textbf{25.52} & \textbf{23.21} & 34.45 & 24.42 & 21.82 & 23.74 \\ \cmidrule{2-10}
 & \multirow{5}{*}{\shortstack[l]{DNF\\+IU}} & MRR & \textbf{28.17} & \textbf{19.34} & \textbf{15.04} & \textbf{30.22} & \textbf{19.69} & \textbf{15.69} & \textbf{18.24} \\
 &  & HIT@1 & \textbf{11.94} & \textbf{9.80} & \textbf{7.80} & 15.92 & \textbf{10.59} & \textbf{8.33} & \textbf{9.70} \\
 &  & HIT@3 & \textbf{37.01} & \textbf{22.08} & \textbf{16.13} & \textbf{37.10} & \textbf{21.94} & \textbf{16.80} & \textbf{20.21} \\
 &  & HIT@10 & \textbf{58.66} & \textbf{38.16} & \textbf{29.35} & \textbf{58.51} & \textbf{38.03} & \textbf{30.39} & \textbf{35.33} \\
 &  & RA-Oracle & \textbf{33.25} & \textbf{25.52} & \textbf{23.21} & \textbf{35.30} & \textbf{25.51} & \textbf{23.07} & \textbf{24.84} \\ \midrule
\multirow{10}{*}{NewLook} & \multirow{5}{*}{\shortstack[l]{DNF\\+IUd}} & MRR & 16.28 & 11.92 & 9.46 & 19.90 & 12.89 & 10.55 & 12.05 \\
 &  & HIT@1 & 3.04 & 4.45 & 3.67 & 7.42 & 5.69 & 4.66 & 5.18 \\
 &  & HIT@3 & 20.00 & 12.89 & 9.59 & 23.76 & 13.74 & 10.85 & 12.79 \\
 &  & HIT@10 & 44.20 & 26.73 & 20.83 & 45.90 & 27.26 & 21.98 & 25.63 \\
 &  & RA-Oracle & 21.08 & 16.65 & 15.73 & 24.30 & 17.30 & 16.34 & 17.18 \\ \cmidrule{2-10}
 & \multirow{5}{*}{\shortstack[l]{DNF\\+IUD}} & MRR & 16.28 & 11.92 & 9.46 & 20.30 & 13.10 & 10.61 & 12.18 \\
 &  & HIT@1 & 3.04 & 4.45 & 3.67 & 7.56 & 5.77 & 4.68 & 5.23 \\
 &  & HIT@3 & 20.00 & 12.89 & 9.59 & 24.38 & 14.00 & 10.91 & 12.96 \\
 &  & HIT@10 & 44.20 & 26.73 & 20.83 & 46.89 & 27.76 & 22.12 & 25.96 \\
 &  & RA-Oracle & 21.08 & 16.65 & 15.73 & 24.74 & 17.56 & 16.43 & 17.35 \\ 
 \bottomrule
\end{tabular}
\end{table}

\begin{table}[t]
\centering
\scriptsize
\caption{Benchmark results(\%) on EPFO queries of NELL.}
\begin{tabular}{lllrrrrrrrrrr}
\toprule
\multirow{2}{*}{\shortstack[l]{CQA\\  Model}} & \multirow{2}{*}{\shortstack[l]{Normal\\ Form}} & \multirow{2}{*}{Metric} & \multicolumn{9}{c}{(\# anchor nodes, max   length of projection chains)} & \multicolumn{1}{c}{\multirow{2}{*}{AVG.}} \\
\cmidrule{4-12}
 &  & \multicolumn{1}{c}{} & \multicolumn{1}{c}{(1,1)} & \multicolumn{1}{c}{(1,2)} & \multicolumn{1}{c}{(1,3)} & \multicolumn{1}{c}{(2,1)} & \multicolumn{1}{c}{(2,2)} & \multicolumn{1}{c}{(2,3)} & \multicolumn{1}{c}{(3,1)} & \multicolumn{1}{c}{(3,2)} & \multicolumn{1}{c}{(3,3)} & \multicolumn{1}{c}{} \\ \midrule
\multirow{15}{*}{BetaE} & \multirow{5}{*}{DM} & MRR & 28.75 & 10.01 & 9.96 & 15.81 & 10.06 & 9.54 & 11.85 & 10.63 & 9.90 & 10.29 \\
 &  & HIT@1 & 20.41 & 5.87 & 6.15 & 10.57 & 6.34 & 6.08 & 7.65 & 7.14 & 6.51 & 6.77 \\
 &  & HIT@3 & 31.79 & 10.53 & 10.28 & 17.44 & 10.69 & 9.98 & 12.98 & 11.28 & 10.50 & 10.93 \\
 &  & HIT@10 & 45.28 & 17.71 & 17.16 & 26.27 & 17.33 & 16.14 & 19.99 & 17.38 & 16.39 & 17.05 \\
 &  & RA-Oracle & 24.59 & 14.63 & 15.42 & 16.46 & 14.04 & 13.66 & 13.70 & 13.80 & 13.36 & 13.64 \\ \cmidrule{2-13}
 & \multirow{5}{*}{\shortstack[l]{DM\\+I}} & MRR & 28.75 & 10.01 & 9.96 & 15.81 & 10.06 & 9.54 & 11.95 & 10.64 & 9.90 & 10.29 \\
 &  & HIT@1 & 20.41 & 5.87 & 6.15 & 10.57 & 6.34 & 6.08 & 7.76 & 7.16 & 6.51 & 6.78 \\
 &  & HIT@3 & 31.79 & 10.53 & 10.28 & 17.44 & 10.69 & 9.98 & 13.15 & 11.29 & 10.50 & 10.94 \\
 &  & HIT@10 & 45.28 & 17.71 & 17.16 & 26.27 & 17.33 & 16.14 & 20.05 & 17.40 & 16.41 & 17.06 \\
 &  & RA-Oracle & 24.59 & 14.63 & 15.42 & 16.46 & 14.04 & 13.66 & 13.82 & 13.82 & 13.36 & 13.65 \\ \cmidrule{2-13}
 & \multirow{5}{*}{\shortstack[l]{DNF\\+IU}} & MRR & 28.75 & 10.01 & 9.96 & 18.48 & 11.90 & 11.06 & 15.41 & 13.54 & 12.35 & 12.73 \\
 &  & HIT@1 & 20.41 & 5.87 & 6.15 & 12.02 & 7.37 & 6.94 & 9.66 & 8.89 & 7.99 & 8.24 \\
 &  & HIT@3 & 31.79 & 10.53 & 10.28 & 20.33 & 12.57 & 11.59 & 16.96 & 14.41 & 13.10 & 13.53 \\
 &  & HIT@10 & 45.28 & 17.71 & 17.16 & 31.41 & 20.85 & 19.00 & 26.67 & 22.59 & 20.77 & 21.45 \\
 &  & RA-Oracle & 24.59 & 14.63 & 15.42 & 21.26 & 17.33 & 16.29 & 19.83 & 18.88 & 17.61 & 17.90 \\ \midrule
\multirow{15}{*}{LogicE} & \multirow{5}{*}{DM} & MRR & \textbf{35.23} & \textbf{13.53} & \textbf{13.75} & 19.42 & 12.66 & 12.04 & 14.99 & 13.08 & 12.24 & 12.75 \\
 &  & HIT@1 & \textbf{25.74} & \textbf{8.62} & \textbf{8.92} & 13.21 & 8.47 & 8.13 & 9.85 & 9.00 & 8.41 & 8.74 \\
 &  & HIT@3 & \textbf{40.27} & \textbf{14.28} & \textbf{14.73} & 21.61 & 13.61 & 12.83 & 16.59 & 14.09 & 13.12 & 13.72 \\
 &  & HIT@10 & \textbf{53.28} & \textbf{23.05} & \textbf{22.80} & 31.69 & 20.64 & 19.40 & 24.98 & 20.83 & 19.46 & 20.36 \\
 &  & RA-Oracle & \textbf{30.69} & \textbf{19.62} & \textbf{20.32} & 21.75 & 17.89 & 16.97 & 18.71 & 17.66 & 16.72 & 17.23 \\ \cmidrule{2-13}
 & \multirow{5}{*}{\shortstack[l]{DM\\+I}} & MRR & \textbf{35.23} & \textbf{13.53} & \textbf{13.75} & 19.42 & 12.66 & 12.04 & 15.03 & 13.09 & 12.25 & 12.76 \\
 &  & HIT@1 & \textbf{25.74} & \textbf{8.62} & \textbf{8.92} & 13.21 & 8.47 & 8.13 & 9.94 & 9.02 & 8.42 & 8.75 \\
 &  & HIT@3 & \textbf{40.27} & \textbf{14.28} & \textbf{14.73} & 21.61 & 13.61 & 12.83 & 16.58 & 14.10 & 13.14 & 13.74 \\
 &  & HIT@10 & \textbf{53.28} & \textbf{23.05} & \textbf{22.80} & 31.69 & 20.64 & 19.40 & 24.97 & 20.83 & 19.47 & 20.36 \\
 &  & RA-Oracle & \textbf{30.69} & \textbf{19.62} & \textbf{20.32} & 21.75 & 17.89 & 16.97 & 18.74 & 17.68 & 16.73 & 17.24 \\ \cmidrule{2-13}
 & \multirow{5}{*}{\shortstack[l]{DNF\\+IU}} & MRR & \textbf{35.23} & \textbf{13.53} & \textbf{13.75} & \textbf{22.48} & \textbf{15.23} & \textbf{14.18} & \textbf{17.91} & \textbf{16.28} & \textbf{15.19} & \textbf{15.63} \\
 &  & HIT@1 & \textbf{25.74} & \textbf{8.62} & \textbf{8.92} & \textbf{14.90} & \textbf{9.88} & \textbf{9.30} & 11.26 & \textbf{10.81} & \textbf{10.13} & \textbf{10.38} \\
 &  & HIT@3 & \textbf{40.27} & \textbf{14.28} & \textbf{14.73} & \textbf{25.02} & \textbf{16.51} & \textbf{15.23} & \textbf{19.84} & \textbf{17.61} & \textbf{16.37} & \textbf{16.90} \\
 &  & HIT@10 & \textbf{53.28} & \textbf{23.05} & \textbf{22.80} & \textbf{37.42} & \textbf{25.50} & \textbf{23.42} & \textbf{30.98} & \textbf{26.78} & \textbf{24.78} & \textbf{25.62} \\
 &  & RA-Oracle & \textbf{30.69} & \textbf{19.62} & \textbf{20.32} & \textbf{26.45} & \textbf{22.08} & \textbf{20.48} & 23.47 & \textbf{22.83} & \textbf{21.50} & \textbf{21.89} \\
 \midrule
\multirow{5}{*}{NewLook} & \multirow{5}{*}{\shortstack[l]{DNF\\+IUD}} & MRR & 33.59 & 12.06 & 11.42 & 22.09 & 13.59 & 12.16 & 17.73 & 14.70 & 13.21 & 13.81 \\
 &  & HIT@1 & 24.78 & 7.12 & 6.79 & 14.66 & 8.62 & 7.69 & \textbf{11.33} & 9.61 & 8.65 & 9.02 \\
 &  & HIT@3 & 37.19 & 12.73 & 12.00 & 24.27 & 14.47 & 12.79 & 19.23 & 15.71 & 13.95 & 14.67 \\
 &  & HIT@10 & 51.89 & 21.72 & 20.31 & 37.14 & 23.13 & 20.64 & 30.46 & 24.53 & 21.92 & 23.03 \\
 &  & RA-Oracle & 29.50 & 18.03 & 17.76 & 26.10 & 20.28 & 18.24 & \textbf{23.87} & 21.10 & 19.28 & 19.89 \\ \bottomrule
\end{tabular}
\end{table}

\begin{table}[t]
\centering
\scriptsize
\caption{Benchmark results(\%) on queries with negation of NELL.}
\begin{tabular}{lllrrrrrrr}
\toprule
\multirow{2}{*}{\shortstack[l]{CQA\\  Model}} & \multirow{2}{*}{\shortstack[l]{Normal\\ Form}} & \multirow{2}{*}{Metric} & \multicolumn{6}{c}{(\# anchor nodes, max   length of projection chains)} & \multicolumn{1}{c}{\multirow{2}{*}{AVG.}} \\ \cmidrule{4-9}
 &  & \multicolumn{1}{c}{} & \multicolumn{1}{c}{(2,1)} & \multicolumn{1}{c}{(2,2)} & \multicolumn{1}{c}{(2,3)} & \multicolumn{1}{c}{(3,1)} & \multicolumn{1}{c}{(3,2)} & \multicolumn{1}{c}{(3,3)} & \multicolumn{1}{c}{} \\ \midrule
\multirow{15}{*}{BetaE} & \multirow{5}{*}{DM} & MRR & 7.56 & 7.30 & 6.57 & 7.49 & 7.67 & 7.47 & 7.53 \\
 &  & HIT@1 & 3.62 & 3.86 & 3.42 & 4.02 & 4.43 & 4.41 & 4.36 \\
 &  & HIT@3 & 7.98 & 7.42 & 6.68 & 7.72 & 7.93 & 7.74 & 7.79 \\
 &  & HIT@10 & 14.68 & 13.77 & 12.30 & 14.12 & 13.69 & 13.18 & 13.44 \\
 &  & RA-Oracle & 9.89 & 10.73 & 10.23 & 9.62 & 10.46 & 10.58 & 10.47 \\ \cmidrule{2-10}
 & \multirow{5}{*}{\shortstack[l]{DM\\+I}} & MRR & 7.56 & 7.30 & 6.57 & 7.48 & 7.69 & 7.49 & 7.54 \\
 &  & HIT@1 & 3.62 & 3.86 & 3.42 & 3.99 & 4.45 & 4.42 & 4.37 \\
 &  & HIT@3 & 7.98 & 7.42 & 6.68 & 7.72 & 7.95 & 7.75 & 7.80 \\
 &  & HIT@10 & 14.68 & 13.77 & 12.30 & 14.13 & 13.72 & 13.20 & 13.46 \\
 &  & RA-Oracle & 9.89 & 10.73 & 10.23 & 9.60 & 10.48 & 10.60 & 10.49 \\ \cmidrule{2-10}
 & \multirow{5}{*}{\shortstack[l]{DNF\\+IU}} & MRR & 7.56 & 7.30 & 6.57 & 8.12 & 8.49 & 8.39 & 8.34 \\
 &  & HIT@1 & 3.62 & 3.86 & 3.42 & 4.16 & 4.82 & 4.85 & 4.74 \\
 &  & HIT@3 & 7.98 & 7.42 & 6.68 & 8.42 & 8.78 & 8.68 & 8.62 \\
 &  & HIT@10 & 14.68 & 13.77 & 12.30 & 15.68 & 15.36 & 15.00 & 15.09 \\
 &  & RA-Oracle & 9.89 & 10.73 & 10.23 & 10.75 & 11.96 & 12.26 & 11.96 \\ \midrule
\multirow{15}{*}{LogicE} & \multirow{5}{*}{DM} & MRR & \textbf{9.62} & \textbf{9.51} & \textbf{9.05} & 9.09 & 9.48 & 9.50 & 9.46 \\
 &  & HIT@1 & \textbf{4.56} & \textbf{5.30} & \textbf{5.47} & 4.74 & 5.64 & 5.99 & 5.74 \\
 &  & HIT@3 & \textbf{10.34} & \textbf{9.98} & \textbf{9.31} & 9.66 & 10.01 & 10.03 & 9.99 \\
 &  & HIT@10 & \textbf{19.20} & \textbf{17.13} & \textbf{15.59} & 17.47 & 16.61 & 15.94 & 16.33 \\
 &  & RA-Oracle & \textbf{13.32} & \textbf{13.98} & \textbf{13.48} & 12.26 & 13.20 & 13.26 & 13.21 \\ \cmidrule{2-10}
 & \multirow{5}{*}{\shortstack[l]{DM\\+I}} & MRR & \textbf{9.62} & \textbf{9.51} & \textbf{9.05} & 9.09 & 9.48 & 9.50 & 9.47 \\
 &  & HIT@1 & \textbf{4.56} & \textbf{5.30} & \textbf{5.47} & 4.76 & 5.65 & 5.99 & 5.75 \\
 &  & HIT@3 & \textbf{10.34} & \textbf{9.98} & \textbf{9.31} & 9.64 & 10.00 & 10.04 & 9.99 \\
 &  & HIT@10 & \textbf{19.20} & \textbf{17.13} & \textbf{15.59} & 17.46 & 16.61 & 15.95 & 16.34 \\
 &  & RA-Oracle & \textbf{13.32} & \textbf{13.98} & \textbf{13.48} & 12.23 & 13.19 & 13.26 & 13.20 \\ \cmidrule{2-10}
 & \multirow{5}{*}{\shortstack[l]{DNF\\+IU}} & MRR & \textbf{9.62} & \textbf{9.51} & \textbf{9.05} & \textbf{9.65} & \textbf{10.42} & \textbf{10.63} & \textbf{10.43} \\
 &  & HIT@1 & \textbf{4.56} & \textbf{5.30} & \textbf{5.47} & \textbf{4.83} & \textbf{6.03} & \textbf{6.51} & \textbf{6.16} \\
 &  & HIT@3 & \textbf{10.34} & \textbf{9.98} & \textbf{9.31} & \textbf{10.19} & \textbf{11.01} & \textbf{11.25} & \textbf{11.02} \\
 &  & HIT@10 & \textbf{19.20} & \textbf{17.13} & \textbf{15.59} & \textbf{19.02} & \textbf{18.63} & \textbf{18.20} & \textbf{18.34} \\
 &  & RA-Oracle & \textbf{13.32} & \textbf{13.98} & \textbf{13.48} & \textbf{13.20} & \textbf{14.68} & \textbf{15.13} & \textbf{14.77} \\ \midrule
\multirow{10}{*}{NewLook} & \multirow{5}{*}{\shortstack[l]{DNF\\+IUd}} & MRR & 5.46 & 5.35 & 5.08 & 5.39 & 5.77 & 5.98 & 5.82 \\
 &  & HIT@1 & 2.45 & 2.62 & 2.40 & 2.33 & 2.93 & 3.10 & 2.96 \\
 &  & HIT@3 & 5.46 & 5.33 & 5.01 & 5.38 & 5.84 & 6.04 & 5.88 \\
 &  & HIT@10 & 11.14 & 10.25 & 9.71 & 11.21 & 10.91 & 11.09 & 10.97 \\
 &  & RA-Oracle & 5.60 & 7.25 & 7.87 & 6.39 & 7.84 & 8.81 & 8.21 \\ \cmidrule{2-10}
 & \multirow{5}{*}{\shortstack[l]{DNF\\+IUD}} & MRR & 5.46 & 5.35 & 5.08 & 5.44 & 5.81 & 5.99 & 5.85 \\
 &  & HIT@1 & 2.45 & 2.62 & 2.40 & 2.35 & 2.94 & 3.11 & 2.97 \\
 &  & HIT@3 & 5.46 & 5.33 & 5.01 & 5.45 & 5.88 & 6.05 & 5.91 \\
 &  & HIT@10 & 11.14 & 10.25 & 9.71 & 11.28 & 11.00 & 11.12 & 11.03 \\
 &  & RA-Oracle & 5.60 & 7.25 & 7.87 & 6.43 & 7.89 & 8.83 & 8.24 \\ \bottomrule
\end{tabular}
\end{table}


\begin{thebibliography}{10}

\bibitem{DBLP:conf/iclr/ArakelyanDMC21}
Erik Arakelyan, Daniel Daza, Pasquale Minervini, and Michael Cochez.
\newblock Complex query answering with neural link predictors.
\newblock In {\em {ICLR}}. OpenReview.net, 2021.

\bibitem{auer2007dbpedia}
S{\"{o}}ren Auer, Christian Bizer, Georgi Kobilarov, Jens Lehmann, Richard
  Cyganiak, and Zachary~G. Ives.
\newblock Dbpedia: {A} nucleus for a web of open data.
\newblock In {\em {ISWC/ASWC}}, volume 4825 of {\em Lecture Notes in Computer
  Science}, pages 722--735. Springer, 2007.

\bibitem{freebase}
Kurt~D. Bollacker, Colin Evans, Praveen Paritosh, Tim Sturge, and Jamie Taylor.
\newblock Freebase: a collaboratively created graph database for structuring
  human knowledge.
\newblock In {\em SIGMOD}, pages 1247--1250, 2008.

\bibitem{DBLP:conf/nips/BordesUGWY13}
Antoine Bordes, Nicolas Usunier, Alberto Garc{\'{\i}}a{-}Dur{\'{a}}n, Jason
  Weston, and Oksana Yakhnenko.
\newblock Translating embeddings for modeling multi-relational data.
\newblock In {\em {NIPS}}, pages 2787--2795, 2013.

\bibitem{carlson2010toward}
Andrew Carlson, Justin Betteridge, Bryan Kisiel, Burr Settles, Estevam
  R.~Hruschka Jr., and Tom~M. Mitchell.
\newblock Toward an architecture for never-ending language learning.
\newblock In {\em {AAAI}}. {AAAI} Press, 2010.

\bibitem{DBLP:conf/www/ChoudharyRKSR21}
Nurendra Choudhary, Nikhil Rao, Sumeet Katariya, Karthik Subbian, and
  Chandan~K. Reddy.
\newblock Self-supervised hyperboloid representations from logical queries over
  knowledge graphs.
\newblock In {\em {WWW}}, pages 1373--1384. {ACM} / {IW3C2}, 2021.

\bibitem{DBLP:conf/iclr/DasDZVDKSM18}
Rajarshi Das, Shehzaad Dhuliawala, Manzil Zaheer, Luke Vilnis, Ishan Durugkar,
  Akshay Krishnamurthy, Alex Smola, and Andrew McCallum.
\newblock Go for a walk and arrive at the answer: Reasoning over paths in
  knowledge bases using reinforcement learning.
\newblock In {\em {ICLR} (Poster)}. OpenReview.net, 2018.

\bibitem{davey2002introduction}
Brian~A Davey and Hilary~A Priestley.
\newblock {\em Introduction to Lattices and Order}.
\newblock Cambridge University Press, 2002.

\bibitem{DBLP:conf/i-semantics/EhrlingerW16}
Lisa Ehrlinger and Wolfram W{\"{o}}{\ss}.
\newblock Towards a definition of knowledge graphs.
\newblock In {\em SEMANTiCS (Posters, Demos, SuCCESS)}, volume 1695 of {\em
  {CEUR} Workshop Proceedings}. CEUR-WS.org, 2016.

\bibitem{DBLP:conf/nips/HamiltonBZJL18}
William~L. Hamilton, Payal Bajaj, Marinka Zitnik, Dan Jurafsky, and Jure
  Leskovec.
\newblock Embedding logical queries on knowledge graphs.
\newblock In {\em NeurIPS}, pages 2030--2041, 2018.

\bibitem{DBLP:conf/aaai/KotnisLN21}
Bhushan Kotnis, Carolin Lawrence, and Mathias Niepert.
\newblock Answering complex queries in knowledge graphs with bidirectional
  sequence encoders.
\newblock In {\em {AAAI}}, pages 4968--4977. {AAAI} Press, 2021.

\bibitem{DBLP:conf/kdd/LiuDJZT21}
Lihui Liu, Boxin Du, Heng Ji, ChengXiang Zhai, and Hanghang Tong.
\newblock Neural-answering logical queries on knowledge graphs.
\newblock In {\em {KDD}}, pages 1087--1097. {ACM}, 2021.

\bibitem{DBLP:journals/corr/abs-2103-00418}
Francois P.~S. Luus, Prithviraj Sen, Pavan Kapanipathi, Ryan Riegel, Ndivhuwo
  Makondo, Thabang Lebese, and Alexander~G. Gray.
\newblock Logic embeddings for complex query answering.
\newblock {\em CoRR}, abs/2103.00418, 2021.

\bibitem{mccarthy1965lisp}
John McCarthy, Michael~I Levin, Paul~W Abrahams, Daniel~J Edwards, and
  Timothy~P Hart.
\newblock {\em LISP 1.5 Programmer's Manual}.
\newblock MIT Press, 1965.

\bibitem{DBLP:conf/iclr/RenHL20}
Hongyu Ren, Weihua Hu, and Jure Leskovec.
\newblock Query2box: Reasoning over knowledge graphs in vector space using box
  embeddings.
\newblock In {\em {ICLR}}. OpenReview.net, 2020.

\bibitem{DBLP:conf/nips/RenL20}
Hongyu Ren and Jure Leskovec.
\newblock Beta embeddings for multi-hop logical reasoning in knowledge graphs.
\newblock In {\em NeurIPS}, 2020.

\bibitem{robinson2001handbook}
Alan~JA Robinson and Andrei Voronkov.
\newblock {\em Handbook of Automated Reasoning}, volume~1.
\newblock Elsevier, 2001.

\bibitem{suchanek2007yago}
Fabian~M. Suchanek, Gjergji Kasneci, and Gerhard Weikum.
\newblock Yago: a core of semantic knowledge.
\newblock In {\em {WWW}}, pages 697--706. {ACM}, 2007.

\bibitem{DBLP:conf/nips/SunAB0C20}
Haitian Sun, Andrew~O. Arnold, Tania Bedrax{-}Weiss, Fernando Pereira, and
  William~W. Cohen.
\newblock Faithful embeddings for knowledge base queries.
\newblock In {\em NeurIPS}, 2020.

\bibitem{toutanova2015observed}
Kristina Toutanova and Danqi Chen.
\newblock Observed versus latent features for knowledge base and text
  inference.
\newblock In {\em ACL Workshop on Continuous Vector Space Models and Their
  Compositionality}, pages 57--66, 2015.

\bibitem{vankov2020training}
Ivan~I Vankov and Jeffrey~S Bowers.
\newblock Training neural networks to encode symbols enables combinatorial
  generalization.
\newblock {\em Philosophical Transactions of the Royal Society B},
  375(1791):20190309, 2020.

\end{thebibliography}
\end{document}